\documentclass[lettersize,journal]{IEEEtran}
\usepackage{amsmath, amsfonts, bm}
\usepackage{algorithmic}
\usepackage{algorithm}
\usepackage{array}
\usepackage[caption=false,font=normalsize,labelfont=sf,textfont=sf]{subfig}
\usepackage{textcomp}
\usepackage{stfloats}
\usepackage{url}
\usepackage{makecell}
\usepackage{multirow}
\usepackage{verbatim}
\usepackage{graphicx}
\usepackage{cite}
\usepackage{svg}
\begin{document}

\title{ConvBLS: An Effective and Efficient Incremental Convolutional Broad Learning System \\for Image Classification}

\author{\IEEEauthorblockN{Chunyu Lei, C. L. Philip Chen\IEEEmembership{, Fellow,~IEEE}, Jifeng Guo, and Tong Zhang\IEEEmembership{, Member,~IEEE}}
    \thanks{This work was funded in part by the National Key Research and Development Program of China under number 2019YFA0706200, in part by the National Natural Science Foundation of China grant under numbers 62076102, 62222603, and 92267203, in part by the Guangdong Natural Science Funds for Distinguished Young Scholar under number 2020B1515020041, and in part by the Program for Guangdong Introducing Innovative and Entrepreneurial Teams (2019ZT08X214).
    \emph{(Corresponding author: Tong Zhang.)}}

    \thanks{Chunyu Lei, C. L. Philip Chen, Jifeng Guo, and Tong Zhang are with the Guangdong Provincial Key Laboratory of Computational Intelligence and Cyberspace Information, the School of Computer Science and Engineering, South China University of Technology, Guangzhou 510006, China, and are with the Brain and Affective Cognitive Research Center, Pazhou Lab, Guangzhou 510335, China (e-mail: cscylei@gmail.com; philipchen@scut.edu.cn; linda.guojf@outlook.com; tony@scut.edu.cn).}
}

\maketitle

\begin{abstract}
Deep learning generally suffers from enormous computational resources and time-consuming training processes. Broad Learning System (BLS) and its convolutional variants have been proposed to mitigate these issues and have achieved superb performance in image classification.
However, the existing convolutional-based broad learning system (C-BLS) either lacks an efficient training method and incremental learning capability or suffers from poor performance. To this end, we propose a convolutional broad learning system (ConvBLS) based on the spherical K-means (SKM) algorithm and two-stage multi-scale (TSMS) feature fusion, which consists of the convolutional feature (CF) layer, convolutional enhancement (CE) layer, TSMS feature fusion layer, and output layer. 
First, unlike the current C-BLS, the simple yet efficient SKM algorithm is utilized to learn the weights of CF layers. Compared with random filters, the SKM algorithm makes the CF layer learn more comprehensive spatial features. Second, similar to the vanilla BLS, CE layers are established to expand the feature space. Third, the TSMS feature fusion layer is proposed to extract more effective multi-scale features through the integration of CF layers and CE layers. 
Thanks to the above design and the pseudo-inverse calculation of the output layer weights, our proposed ConvBLS method is unprecedentedly efficient and effective. Finally, the corresponding incremental learning algorithms are presented for rapid remodeling if the model deems to expand. Experiments and comparisons demonstrate the superiority of our method.
\end{abstract}


\section{Introduction}
\IEEEPARstart{W}{ith} the rapid development of deep learning, extensive breakthrough achievements have been acquired in various tasks, including image classification\cite{lecun1989handwritten, krizhevsky2012imagenet,simonyan2014very, szegedy2015going, he2016deep, huang2017densely}, semantic segmentation\cite{long2015fully, chen2017deeplab, lin2017refinenet}, object detection\cite{girshick2015fast, ren2015faster, redmon2016you}, etc. Unfortunately, deep neural networks (DNNs) with enormous parameters generally suffer from the time-consuming training processes due to over-complicated architectures. Additionally, once the network structure needs to be modified to obtain better accuracy, a complete retraining process is inevitable. 

To overcome the aforementioned weaknesses, BLS has been proposed as an alternative model to deep learning algorithms\cite{chen2017broad,chen2018universal}. Owing to its effectiveness and efficiency, BLS has attracted increasing attention and has been frequently used in many different fields\cite{gong2021research}. Nevertheless, as a specific flattened fully-connected neural network, the vanilla BLS handles image data in an unrolled one-dimensional vector manner, which makes it challenging to extract hierarchical spatial features\cite{chen2018universal}. Importantly, superior spatial image features are essential for performance improvement in image classification tasks\cite{lecun1989handwritten, liu2017broad, liu2021multi}.

For the purpose of improving the spatial feature representation ability of BLS, a mass of C-BLS variants\cite{chen2018universal, yang2018cnn, yu2019broad, li2019cnn, chen2020cnn, tang2020combining, ma2021multiscale} have been proposed by introducing local inductive bias of convolution operation. Some works\cite{chen2020cnn, tang2020combining} immediately feed the final features of the last layer of CNNs into the original BLS. Other works\cite{chen2018universal, yang2018cnn, yu2019broad, li2019cnn} cascade feature layers and enhancement layers to improve the final classification performance. Despite these progresses, they still lack flexible and rich multi-scale features for obtaining excellent performance \cite{grauman2005pyramid, lazebnik2006beyond}.

Furthermore, the existing optimization methods of convolutional filters for C-BLS are divided into two categories, including random convolutional filters (RCF)\cite{chen2018universal, yang2018cnn, yu2019broad, chen2020cnn, tang2020combining, ma2021multiscale} and trained convolutional filters (TCF) via a gradient descent algorithm \cite{li2019cnn}. For RCF-based methods, the weights of convolutional filters are randomly sampled under a given distribution. They can boost the performance to some extent but still suffer from the following problems: 1) the model stability is poor, and 2) there remains a tremendous performance gap between them and typical DNNs. For TCF-based methods, the weights of convolutional filters are optimized by a stochastic gradient descent algorithm using back-propagation (BP). 
As a result, the shortcomings of deep learning, such as incredible training time, massive computing resource consumption, and poor generalization, also exist in these methods.

Last but not least, the incremental learning capability makes BLS dynamically adjust the network structure without a tedious retraining process. However, a majority of existing C-BLS variants \cite{chen2018universal, yang2018cnn, li2019cnn, tang2020combining, ma2021multiscale} fail to equipped with incremental learning ability except BCNN\cite{yu2019broad} and CNNBL\cite{chen2020cnn}. Even so, both BCNN\cite{yu2019broad} and CNNBL\cite{chen2020cnn} are optimized by RCF-based methods and thus can not achieve the required performance. Therefore, it is significant to design incremental learning algorithms for models whose convolutional filters require training.

Considering the above issues, we propose an effective and efficient convolutional broad learning system (ConvBLS) based on the SKM algorithm and TSMS feature fusion. It adopts a naive unsupervised learning algorithm, SKM, for the efficient filter learning of CF layers, which only requires low computing resources and training time. Additionally, the orthogonal CE layer is designed to expand the feature space. To further mine the features of CF layers and CE layers, a TSMS feature fusion layer is proposed to obtain abundant multi-scale features used for decision. Importantly, because of more discriminative features extracted by CF and CE layers, ConvBLS is naturally suitable for semi-supervised learning scenarios with few labeled data. Lastly, it is equipped with two proposed incremental learning algorithms to achieve fast remodeling without the tedious retraining process. The main contributions of this article are summarized as follows.
\begin{enumerate}
    \item A novel and effective ConvBLS architecture is developed, which is composed of the CF layer, CE layer, TSMS feature fusion layer, and output layer. Among them, the TSMS feature fusion layer is designed for the first time to extract richer multi-scale features by combining CF layers, CE layers, and SPP techniques.
    \item We present a rapid and efficient training method for ConvBLS. Due to the powerful unsupervised feature extraction capability, our method can be adapted to semi-supervised learning tasks without modification.
    \item We design two incremental learning algorithms to adjust the model dynamically. To our best knowledge, it is the first time to propose incremental learning algorithms for C-BLS methods whose convolutional filters need to be trained.
\end{enumerate}

The rest of this paper is organized as follows. In Section II, the related works of this article are given. Section III illustrates the technical details of the proposed ConvBLS. Extensive experiments are carried out to demonstrate the effectiveness and efficiency of our method in Section IV. Finally, Section V concludes this article and discusses several future research directions. 

\section{Related Works}
The main topic of this article is to design a valid ConvBLS architecture and develop a rapid training algorithm and corresponding incremental learning algorithms. Consequently, in this Section, BLS and its convolutional variants are first reviewed. Then, to understand the TSMS feature fusion, the typical SPP technique is recalled. Finally, the existing convolutional filter training methods without supervised signals are introduced to improve the training method of the C-BLS.

\subsection{Broad Learning System and Its Convolutional Variants}
BLS\cite{chen2017broad, chen2018universal} is an alternative model to DNNs, that consists of three parts, feature nodes, enhancement nodes, and output nodes. First, the input data is randomly mapped into feature nodes. Then all feature nodes are randomly mapped into enhancement nodes. After that, all feature nodes and enhancement nodes are connected with the output nodes. Because the weights of the output nodes can be obtained by pseudo-inverse calculation, the training of BLS is extremely fast. More details about BLS and corresponding incremental learning algorithms can be found in \cite{chen2017broad}.

Different from the flat single hidden layer architecture of BLS, several variants\cite{chen2018universal} such as CFBLS, CEBLS, and CFEBLS have been developed to elevate the performance by deepening the network. However, the performance of these variants is poor on image classification tasks due to a lack of inductive bias and poor feature extraction ability. To tackle these problems, Chen et al. \cite{chen2018universal} presented CCFBLS, a pioneer of C-BLS, to extract image features using the random convolutional layer. Unlike CCFBLS, Yang et al. \cite{yang2018cnn} leveraged principal component analysis to reduce the dimension of features extracted from random convolutional layers, which reduces the model complexity. Yu et al. \cite{yu2019broad} proposed BCNN and related incremental learning algorithms, which have excellent results in fault diagnosis. Similarly, CNNBL\cite{chen2020cnn}, RCNNBL\cite{tang2020combining}, and MRC-BLS\cite{ma2021multiscale}, characterized by random convolutional filters, have been successfully applied to facial expression recognition, music classification, and hyperspectral image classification tasks, respectively. In addition, instead of random convolutional filters, Li et al.\cite{li2019cnn} improved the performance in image classification by utilizing the Adam algorithm\cite{DBLP:journals/corr/KingmaB14} to fine-tune the weights of convolutional layers. 

Despite these advances, they still have their own problems. For example, the RCF-based methods\cite{chen2018universal, yang2018cnn, yu2019broad, chen2020cnn, tang2020combining, ma2021multiscale} lack strong feature extraction ability. Conversely, the TCF-based methods\cite{li2019cnn} require tedious training time inherited from BP algorithms. Thus, in this study, we try to synthesize the merits of both methods to design an efficient and effective ConvBLS.

\subsection{Spatial Pyramid Pooling}
Being one of the most successful techniques in the traditional computer vision community, spatial pyramid matching\cite{grauman2005pyramid, lazebnik2006beyond} (a.k.a. spatial pyramid pooling, SPP) has been suggested to extract more robust multi-scale feature representations. Inspired by \cite{grauman2005pyramid, lazebnik2006beyond}, SPP-net\cite{he2015spatial} has been proposed by He et al., which first couples SPP into CNNs for image classification and object detection tasks. Specifically, SPP pools the input feature maps through multiple branches and combines the output features of all these branches to form the final spatial pyramid features. To obtain semantic features at different scales, pooling operators in different branches, with various pooling window sizes and strides, are performed separately. Among them, the number of layers of a pyramid (i.e., the number of branches) and the number of features (i.e., the number of spatial bins) in each layer of a pyramid usually need to be manually specified. Despite its conceptual simplicity, SPP is more efficient than the approaches that use more complex spatial paradigms.
After that, SPP has been widely applied to various recognition tasks such as hyperspectral image classification\cite{yue2016deep}, hand gesture recognition\cite{tan2021convolutional, ashiquzzaman2020compact}, and traffic sign recognition\cite{dewi2020evaluation, tai2020deep}. Unlike existing work that explores SPP in CNNs for specific tasks, we propose to combine the concept of SPP and BLS to extract more effective and comprehensive TSMS features.

\subsection{Training Convolutional Filters without Supervised Signals}  
As the successful application of CNNs in computer vision\cite{lecun1989handwritten, krizhevsky2012imagenet, simonyan2014very}, the optimization techniques of convolutional filters fall into two categories. The first category refers to the methods that optimize convolutional filters through the pseudo-label and BP algorithm. For example, the generative adversarial networks and their variants\cite{goodfellow2020generative, radford2015unsupervised, chen2016infogan} train a generator and a discriminator by the learning strategy as the rule of the minimax game. Among them, labels used for model training can be obtained easily by the program itself. DeepCluster\cite{caron2018deep} iteratively clusters the sample features with K-means and uses the generated assignments as supervision to update convolutional filters. Similarly, Exemplar-CNN\cite{dosovitskiy2014discriminative} and SimCLR\cite{chen2020simple} generate pseudo-labels by image transformations and use them for filter training. Despite their superb performance, they require comparable or even 
larger training costs than typical supervised DNNs.

The second utilizes simple unsupervised learning algorithms to optimize convolutional filters without the BP algorithm. Coates et al.\cite{coates2011analysis, coates2012learning, coates2011selecting} used the K-means algorithm to train filters for extracting features on input image maps in a convolutional manner. With an elaborate feature coding scheme, these methods exceeded many complex unsupervised learning methods in image classification. However, compared to typical CNNs\cite{krizhevsky2012imagenet, szegedy2015going, he2016deep, huang2017densely}, these models have enormous convolutional filters (e.g., 4096) and few convolutional layers, which require huge running memory and lack hierarchical spatial features. Most related to ours is the method proposed by Culurciello et al.\cite{culurciello2013analysis}, which utilizes a few filters trained by K-means followed by fully-connected layers for classification in real-time scenarios. The main difference between \cite{culurciello2013analysis} and our work is that we use the CE layer and the TSMS feature fusion layer to leverage features extracted by K-means instead of naive downsampling of inputs as a supplement to final features. Similarly, Dundar et al.\cite{dundar2015convolutional} used convolution k-means to learn filters and supervised methods to learn the connection weights between layers, which not only increases the computation cost but also makes the training pipeline exceedingly complicated. Hence, it is ongoing work to design an elegant algorithm to balance its generalization performance and computational complexity.

\section{Methodology}
The key to developing an effective and efficient model is closely related to the model architecture and training algorithm. Hence, in this Section, the proposed effective ConvBLS architecture is detailed first. Additionally, the simple yet efficient training algorithm for ConvBLS is given. At last, incremental learning algorithms are introduced to avoid tedious retraining processes if the model deems to expand.

\subsection{Convolutional Broad Learning System}
\begin{figure*}[!t]
\centering
\includegraphics[width=6.5in]{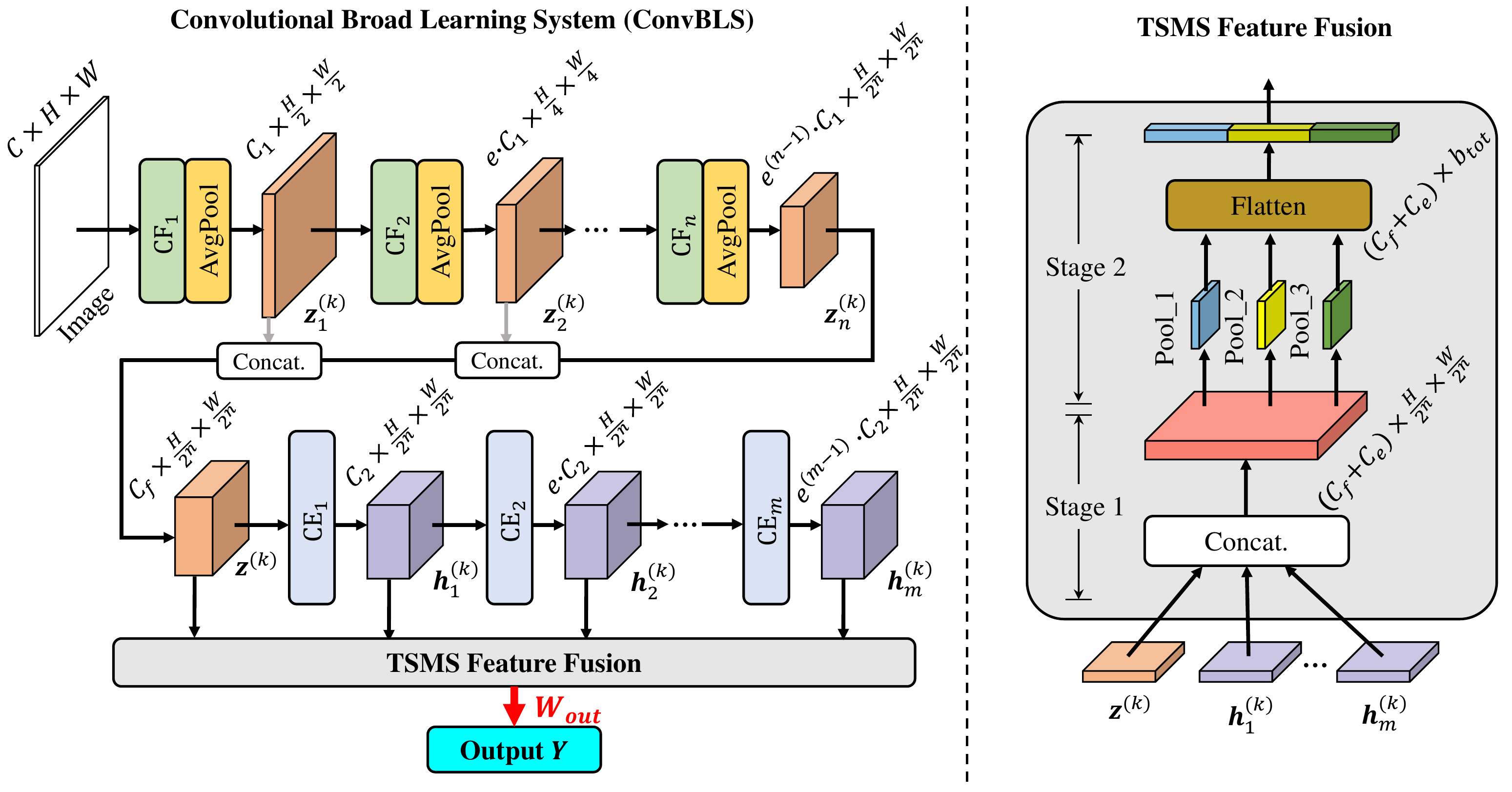}
\caption{\textbf{Topology of Convolutional Broad Learning System.} The proposed ConvBLS stacks $\bm{n}$ CF layers and $\bm{m}$ CE layers one after another for obtaining feature node groups and enhancement node groups, respectively. Among them, all feature node groups are concatenated as the input of the first CE layer. (The gray arrow indicates that appropriate downsampling is needed to ensure successful feature concatenation.) Then, all feature node groups and enhancement node groups are connected to the TSMS feature fusion layer to yield more robust and comprehensive representations. Lastly, the weights matrix $\bm{W}_{out}$ of the output layer is obtained by the ridge regression algorithm. Note that each CF layer is followed by an average pooling layer to reduce feature dimension.
}
\label{framework}
\end{figure*}

For intuitional understanding, the overview architecture of ConvBLS is depicted in Fig. \ref{framework}. 
The proposed ConvBLS is composed of $n$ CF layers denoted as CF$_i$ ($i=1,2,...,n$), $m$ CE layers denoted as CE$_j$ ($j=1,2,...,m$), one TSMS feature fusion layer and one output layer $\bm{Y}$. 
To avoid overfitting, each CF layer follows by an average pooling layer.
Moreover, $C_1$, $C_f$, $C_2$, and $C_e$ are the number of output channels in the CF$_1$ layer, the total number of feature maps in all CF layers, the number of output channels in the CE$_1$ layer, and the total number of feature maps in all CE layers, respectively. Note that a feature map is regarded as a feature node or enhancement node for ConvBLS. Therefore, $C_f$ and $C_e$ also denote the total number of feature and enhancement nodes, respectively. Again, $e$ represents the expansion ratio of the number of output channels throughout the CF and CE layers. Among, $C_2$ is determined by $C_1$, $e$, and $n$. At last, $b_{tot}$ denotes the feature dimension after 
the TSMS feature fusion for each feature map. With $n$ CF layers and $m$ CE layers, $C_1$ and $e$ defined by the user denote the width of the ConvBLS. In the following, we will introduce them in detail.

\subsubsection{Convolutional Feature Layer}
Suppose that the input data set $\bm{X} \in \mathbb{R}^{N \times C \times H \times W}$, where $N$, $C$, $H$, and $W$ denote the number of samples, the number of channels, height, and width, respectively. Taking the $k$th image $\bm{x}^{(k)} \in \mathbb{R}^{C \times H \times W}$ as the input, the output of CF$_i$ (i.e., the $i$th feature node group) can be defined as follows:
\begin{equation}
\label{ConvBLS:z_i^{(k)}}
    \bm{z}_{i}^{(k)} = \bm{\phi} (GConv(\bm{z}_{i-1}^{(k)}; \{ \bm{W}_{f_i}, \bm{\beta}_{f_i}, c_i \})), i=1,2,...,n
\end{equation}
where $\bm{W}_{f_i}$ and $\bm{\beta}_{f_i}$ are the weight matrix and bias matrix, respectively. Among them, $\bm{W}_{f_i}$ is obtained by the SKM algorithm. Similar to \cite{huang2015local}, we set $\bm{\beta}_{f_i}$ to be a zero matrix to simplify the learning of CF layers. Because it is difficult for the SKM algorithm to learn features in a very high-dimensional feature space, we split the high-dimensional feature space into several sub-spaces and utilize the SKM algorithm to learn filters separately in each subspace. Thus, to divide the subspace and ensure that the filters are only used to extract features in the subspace where they are trained, we use group convolution here represented as $GConv(\cdot)$. In other words, the features from different feature sub-spaces (i.e., different feature groups) do not interact. And $c_i$ is the number of feature maps for each feature group in the CF$_i$ layer. Particularly, we set $c_1$ to be the number of channels of the input image for the CF$_1$ layer. As for the subsequent CF layers, $c_i$ and $c_r$ can be different for $i\ne r$. Without loss of generality, we have $c_i = c_r$, where $i, r \in \{2,3,...,n\}$. 
At last, $\bm{\phi} (\cdot)$ represents the ReLU activation function. In other words, each CF layer utilizes the output of its precursor CF layer to obtain more abstract features. Furthermore, in the CF$_1$ layer, the input $\bm{z}_{0}^{(k)}$ is defined as $\bm{x}^{(k)}$. Unlike the group convolution typically used in DNNs, we pre-process all patches extracted from previous feature maps by normalization and whitening before convolution.

\subsubsection{Convolutional Enhancement Layer}
Denote $\bm{z}^{(k)}\equiv [\bm{z}_1^{(k)},\bm{z}_2^{(k)},...,\bm{z}_n^{(k)}]$, which is the concatenation of all feature node groups. To keep the spatial size of the feature nodes consistent, pooling operations are used in appropriate locations. Since each feature node is a feature map obtained by convolution, there are still significant spatial relationships amongst the feature values within each map. Consequently, we use convolution for two-dimensional feature enhancement instead of the one-dimensional feature enhancement used in BLS to preserve the spatial relationships of feature values. The output of CE$_j$ (i.e., the $j$th enhancement node group) can be formulated as follows:
\begin{equation}
\label{ConvBLS:h_j^{(k)}}
    \bm{h}_{j}^{(k)} = \bm{\varphi} (Conv(\bm{h}_{j-1}^{(k)}; \{ \bm{W}_{e_j}, \bm{\beta}_{e_j} \})), j=1,2,...,m
\end{equation}
where $\bm{W}_{e_j}$ and $\bm{\beta}_{e_j}$ are randomly generated weight matrix and bias matrix, respectively. $Conv(\cdot)$ stands for convolutional operation and $\bm{\varphi}(\cdot)$ denotes a selected activation function. Similarly, $\bm{h}_0^{(k)}$ is equivalent to $\bm{z}^{(k)}$ in the CE$_1$ layer.

\subsubsection{Two-stage Multi-scale Feature Fusion}
To obtain promising performance, all feature node groups and enhancement node groups (with various feature scales) are concatenated directly to yield the first stage multi-scale features, which can be expressed as follows:
\begin{equation}
\label{TSMS:a_s1^{(k)}}
    \bm{a}^{(k)}_{s_1} = \bm{\psi} (\bm{z}_1^{(k)},\bm{z}_2^{(k)},...,\bm{z}_n^{(k)}, \bm{h}_1^{(k)},\bm{h}_2^{(k)},...,\bm{h}_m^{(k)})
\end{equation}
where $\bm{\psi}(\cdot)$ is a concatenation function and $s_1$ represents the first stage multi-scale features. Subsequently, more reasonable and comprehensive two-stage multi-scale features are attained by the typical SPP technique. The second stage multi-scale features are defined as follows:
\begin{equation}
    \label{TSMS:a_s2^{(k)}}
    \bm{a}^{(k)}_{s_2}=\bm{\tau}(\bm{a}^{(k)}_{s_1}; \{b_1, b_2, ..., b_d\})
\end{equation}
where $d$ is the number of feature pyramid layers and $b_l$ denotes the size of the feature maps for the $l$th layer feature pyramid. $\bm{\tau}(\cdot)$ represents a function combination of SPP and flattening. Suppose that the shape of $a^{(k)}_{s_1}$ is $(C_f +C_e) \times W_{s_1} \times H_{s_1}$, in which $W_{s_1}$ and $H_{s_1}$ are usually equal, thus we set $W_{s_1} = H_{s_1} = v$. Take the $l$th layer pyramid as an example, to obtain the features with the specified scale, we have a pooling layer with the window size $ win = \lceil v/b_l \rceil $ and stride $str = \lfloor v/b_l \rfloor$, where $\lceil\cdot\rceil$ and $\lfloor\cdot\rfloor$ represent ceiling and floor operations. Besides, $s_2$ denotes the second stage multi-scale features. Combining Eq. (\ref{TSMS:a_s1^{(k)}})
and Eq. (\ref{TSMS:a_s2^{(k)}}), the TSMS features $\bm{a}^{(k)}_{s_2}$ of the $k$th sample can be obtained. For simplicity, the subscript $s_2$ is omitted in the rest of this paper.

\subsubsection{Output Layer}
The final representation of each sample is a comprehensive feature vector, which is, to some extent, translation invariant and scales invariant inherited from convolutions and has rich multi-scale semantic information. Similar to the vanilla BLS, the output layer of ConvBLS is still a plain linear classification layer that can be represented as the equation of the form:
\begin{equation}
    \label{OUTPUT:y^{(k)}}
    \bm{y}^{(k)}=\bm{a}^{(k)}\bm{W}_{out}
\end{equation}
where $\bm{W}_{out}$ is the weight matrix and $\bm{y}^{(k)}$ denotes the final output of ConvBLS for input image $\bm{x}^{(k)}$. Furthermore, the outstanding classification performance of ConvBLS can also be ensured by the efficient training algorithm, which will be illustrated later. 

\subsection{Efficient Training Algorithm for ConvBLS}
The training process of ConvBLS is roughly divided into two phases: the first phase is to train the parameters of CF layers using an unsupervised learning algorithm. Conversely, the second phase is to train the parameters of the output layer (i.e., the classifier) using a supervised learning algorithm.

\subsubsection{Unsupervised Learning for the CF Layer}
The weights of convolutional filters for CF layers are trained using the SKM algorithm in a greedy manner. When training the CF$_i$ layer, the extracted features for all training samples at the CF$_{i-1}$ layer are available. We collect the features of all training samples in the CF$_{i-1}$ layer as $\bm{Z}_{i-1}\in \mathbb{R}^{N \times e^{(i-2)}C_{1} \times \frac{H}{2^{i-1}} \times \frac{W}{2^{i-1}}}$. The equation for $\bm{Z}_{i-1}$ is denoted as
$\bm{Z}_{i-1}=[\bm{z}_{i-1}^{(1)},\bm{z}_{i-1}^{(2)},...,\bm{z}_{i-1}^{(N)}], i=1,2,...n$
, where $\bm{Z}$ represents the features of all samples and $\bm{z}$ denotes the features of one sample. Moreover, when the CF$_1$ layer is trained, $\bm{Z}_0$ is defined as $\bm{X}$. Given the number of feature maps $\bm{c}_i$ within each feature group for the CF$_i$ layer, the features extracted from the CF$_{i-1}$ layer can be divided into $\lceil e^{(i-2)}C_{1}/{\bm{c}_i} \rceil$ groups. Additionally, for $\bm{Z}_{i-1,q} \in \mathbb{R}^{N \times c_i \times \frac{H}{2^{i-1}} \times \frac{W}{2^{i-1}}}$, where $q$ represents the $q$th feature group, the training procedure begins by extracting random patches from the feature group $\bm{Z}_{i-1,q}$. Each patch has dimension $w$-by-$w$-by-$c_i$, with $w$ referred to as the receptive field size. After that, pre-processing operations are necessary. First, every patch is normalized by subtracting the mean and dividing by the standard deviation of its elements to normalize the brightness and contrast. 
Subsequently, to overcome the correlations between adjacent pixels, the ZCA whitening transform\cite{ranzato2010factored} should be used. 
We then gather all of the patches and construct a new dataset for the training of this set of convolutional filters. The dataset is represented as $\bm{P}=\{\bm{p}^{(1)},\bm{p}^{(2)},...,\bm{p}^{(S)}\}$, 
where $\bm{p}^{(s)} \in \mathbb{R}^{w \times w \times c_i}$, and $S$ represents the number of patches.

After pre-processing, the core issue becomes how to learn the weights of convolution filters from patches. As we know, the principle of deep neural networks essentially involves a template-matching problem. For CNNs, each convolution filter is a template (i.e., pattern) used to extract the corresponding feature from the response of the precursor layer. Thus, a set of excellent templates must produce similar responses on the same class of samples and vice versa, which is crucial for the classification task. However, the RCF-based methods\cite{chen2018universal, yang2018cnn, yu2019broad, chen2020cnn, tang2020combining, ma2021multiscale} whose convolutional filters are generated randomly can not meet the above requirements. Inspired by the earlier works that use the K-means algorithm for unsupervised feature learning\cite{coates2011analysis, coates2012learning, coates2011selecting, culurciello2013analysis, dundar2015convolutional}, we utilize the  
SKM to learn better convolutional filters (i.e., templates) for CF layers. 
In this context, the data points to be clustered are randomly extracted patches, and the centroids are the convolutional filters used to extract features from the corresponding output of the predecessor CF layer. The algorithm finds the convolutional filters as follows:
\begin{equation}
\label{k-means}
\begin{aligned}
    &\bm{u}^{(s)}_{q,t}:=\begin{cases}
    \bm{w} _{f_i,q,t}^{\top}\bm{p}^{(s)} , &{\text{if}}\ {j == \underset {l} {\text{arg max}} \left | \bm{w} _{f_i,q,l}\bm{p}^{(s)} \right | } \\
    {0,}&{\text{otherwise.}}
    \end{cases} \forall s,t\\
    &\bm{W} _{f_i,q}:=\bm{P}\bm{U}_q ^{\top}+\bm{W} _{f_i,q}\\
    &\bm{w} _{f_i,q,t}:=\frac{\bm{w} _{f_i,q,t}}{\left \| \bm{w} _{f_i,q,t} \right \|}, \forall t
\end{aligned}
\end{equation}
where $\bm{u}^{(s)}_q$ is the code vector associated with the input $\bm{p}^{(s)}$, and $\bm{w} _{f_i,q,t}$ is the $t$-th convolutional filter in CF$_i$ for the $q$th feature group $\bm{Z}_{i-1,q}$. Note that the convolution filters learned in the $q$th feature group can only be utilized to extract more abstract features in the $q$th feature group eventually. Recalling the feature grouping method mentioned earlier, we can attain the weights of convolutional filters associated with other feature groups in a similar manner. Finally, the convolutional filters for all CF layers can be fine-tuned in this way.

\subsubsection{Supervised Learning for the Ouput Layer}
The weights of the output layer are trained using the ridge regression algorithm. To this end, the TSMS features of all training samples that are robust to object deformation are first calculated using Eq. \eqref{TSMS:a_s2^{(k)}}. Denoting the final TSMS feature matrix as $\bm{A}$ and the real label vector as $\hat{\bm{Y}}$, the optimization problem of the output layer is expressed as follows:
\begin{equation}
    \label{Problem_to_be_optimized}
    \underset {\bm{W}_{out}} {\text{arg min}}: \left \| \bm{AW}_{out} - \bm{\hat{Y}} \right \|_{2}^{2} + \lambda \left \| \bm{W}_{out} \right \|_{2}^{2} 
\end{equation}
where $\lambda$ represents the regularization coefficient to balance the mean squared error term and L2 normalization term. Also, the problem is convex, and the solution can be obtained by the ridge regression theory, which produces an approximation to the Moore-Penrose generalized inverse by adding a positive number to the diagonal of $\bm{A}^{\top}\bm{A}$ or $\bm{A}\bm{A}^{\top}$. Therefore, the weights of output layer $\bm{W}_{out}$ can be calculated as follows:
\begin{equation}
    \label{moore_pseudo_inverse}
    \bm{W}_{out}=(\bm{A}^{\top}\bm{A}+\lambda \bm{I})^{-1}A^{\top}\hat{\bm{Y}}
\end{equation}
where $I$ is an identity matrix with the same shape as $\bm{A}^{\top}\bm{A}$.

By combining the above two phases, all the weights to be trained in our ConvBLS can be optimized. It should be noted that since the weights of CF layers are optimized without tedious fine-tuned processes using the BP algorithm, the training procedure of the entire model is extremely efficient. Thanks to the powerful feature extraction capability of the CF layer and CE layer, the training of the output layer can achieve excellent classification performance without requiring a lot of labeled data. Thus, our method also has remarkable performance in semi-supervised classification.

\subsection{Incremental Learning Algorithms}
Incremental learning capability is significant for practical application. However, the existing TCF-based approaches that usually mean higher performance failed to be equipped with incremental learning algorithms. Therefore, two incremental learning algorithms for ConvBLS are developed, i.e., the increment of additional feature nodes and the increment of additional enhancement nodes. Note that due to the unique design of ConvBLS, the input data increment algorithm is similar to that of BLS. Hence, it has been omitted.

As described in Section III.A, the feature nodes and enhancement nodes for ConvBLS are feature maps instead of feature values in BLS. And the number of feature nodes or enhancement nodes for each layer depends on the number of output nodes for the predecessor layer and the corresponding expansion ratio. Therefore, in the following elaboration, we only need to add a certain number of nodes to the CF$_1$ layer or the CE$_1$ layer, and the number of nodes in the rest of the layers will vary according to the predefined expansion ratio.

\subsubsection{Incremental of Additional Enhancement Nodes}
Recall that the weights and biases of CE layers do not require training. Therefore, we add additional enhancement nodes for CE layers to improve the performance quickly. For convenience to introduce the incremental learning algorithm of the enhancement nodes, we rewrite the CE layer with four-dimensional tensors. In other words, the entire training set $\bm{X}$ rather than a single image $\bm{x}^{(k)}$ are fed into the ConvBLS at once to calculate the enhancement features $\bm{H}$. Next, we detail the broad expansion method for adding $C^a_e$ additional enhancement nodes in the CE$_1$ layer. The output of the CE$_j$ layer for additional enhancement nodes can be formulated as follows:
\begin{multline}
\label{incre:H_j}
    \bm{H}_{j}^{C_2+1} = \bm{\varphi} (Conv(\bm{H}_{j-1}^{C_2+1}; \{ \bm{W}_{e_j}^{C_2+1}, \bm{\beta}_{e_j}^{C_2+1} \})),\\
    j=1,2,...m
\end{multline}
where
\begin{equation}
\begin{cases}
\bm{W}_{e_1}^{C_2+1} \in \mathbb{R}^{C_e^a \times C_f \times w \times w}, \quad &j=1\\
\bm{W}_{e_j}^{C_2+1} \in \mathbb{R}^{e^{j-1}C_e^a \times e^{j-2}C_e^a \times w \times w}, \quad &j=2,3,...,m\\
\end{cases}
\end{equation}
and $\bm{\beta}_{e_j}^{C_2+1} \in \mathbb{R}^{e^{j-1}C_e^a}$ are randomly generated. Specifically, similarly to Eq. \eqref{ConvBLS:h_j^{(k)}}, $\bm{H}_{0}^{C_2+1}$ is defined as $\bm{Z}$. And then, the new TSMS features can be formulated as follows:
\begin{equation}
    \label{incre:A_new}
    \bm{A}^{C_2+1}= [\bm{A}^{C_2}|\bm{\sigma}(\bm{H}_{1}^{C_2+1},\bm{H}_{2}^{C_2+1},...,\bm{H}_{m}^{C_2+1})]
\end{equation}
where $\bm{\sigma}(\cdot)$ is a function combination of $\bm{\psi}(\cdot)$ and $\bm{\tau}(\cdot)$ in Eq. \eqref{TSMS:a_s1^{(k)}} and Eq. \eqref{TSMS:a_s2^{(k)}}. Then, we deduce the pseudoinverse of the new matrix as
\begin{equation}
    \label{incre_pi}
    (\bm{A}^{C_2+1})^{+}=\begin{bmatrix}
    (\bm{A}^{C_2})^{+}-\bm{DB}\\
    \bm{B}^{\top}
    \end{bmatrix}
\end{equation}
where $\bm{D}=(\bm{A}^{C_2})^{+}\bm{\sigma}(\bm{H}_{1}^{C_2+1},\bm{H}_{2}^{C_2+1},...,\bm{H}_{m}^{C_2+1})$,
\begin{equation}
    \label{incre:BT}
    \bm{B}^{\top}=
    \begin{cases}
    (\bm{Q})^{+} &{\text{if}\ \bm{Q} \ne \bm{0}}\\
    (1+\bm{D}^{\top}\bm{D})^{-1}\bm{D}^{\top}(\bm{A}^{C_2})^{+} &{\text{if}\ \bm{Q}=\bm{0}}\\
    \end{cases}
\end{equation}
and $\bm{Q}=\bm{\sigma}(\bm{H}_{1}^{C_2+1},\bm{H}_{2}^{C_2+1},...,\bm{H}_{m}^{C_2+1})-\bm{A}^{C_2} \bm{D}$

Again, the new weight matrix is 
\begin{equation}
    \label{incre:W_out}
    \bm{W}_{out}^{C_2+1}=
    \begin{bmatrix}
    \bm{W}_{out}^{C_2}-\bm{DB}^{\top}\hat{\bm{Y}}\\
    \bm{B}^{\top}\hat{\bm{Y}}\\
    \end{bmatrix}
\end{equation}

The incremental learning algorithm of additional enhancement nodes is listed in Algorithm \ref{alg:EnhancementIncremental}.

\subsubsection{Incremental of Additional Feature Nodes}
In some cases, due to the insufficient feature nodes, adding enhancement nodes solely does not meet the performance requirements. Here, we describe the incremental learning for newly incremental feature nodes. Similarly, we rewrite the CF layer in four-dimensional tensors. Assume that there are $C_1$ and $C_2$ output channels in the CF$_1$ and CE$_1$ layers of the initial architecture, respectively. Consider adding $C^a_f$ feature nodes to the CF$_1$ layer, the additional output of the $i$th feature layer can be expressed as follows:
\begin{multline}
\label{FeaIncre:Z_i}
    \bm{Z}_{i}^{C_1+1} = \bm{\phi} (GConv(\bm{Z}_{i-1}^{C_1+1}; \{ \bm{W}_{f_i}^{C_1+1}, \bm{\beta}_{f_i}^{C_1+1}, c_i \})), \\
    i=1,2,...,n
\end{multline}
\begin{algorithm}[H]
\renewcommand{\algorithmicrequire}{\textbf{Input:}}
\renewcommand{\algorithmicensure}{\textbf{Output:}}
\caption{ConvBLS: Increment of $C^a_e$ Additional Enhancement Nodes in the CE$_1$ Layer}
\label{alg:EnhancementIncremental}
\begin{algorithmic}[1]
\REQUIRE training samples $\{\textbf{x}^{(k)}, \hat{\textbf{y}}^{(k)}\}_{k=1}^{N}$.
\ENSURE $\{\textbf{W}_{f_i}\}_{i=1}^{n}$, $\textbf{W}_{out}$.
\FOR{$i=1$ to $n$}
\STATE Group feature maps according to $c_i$. For each feature group, execute patches extraction and preprocessing, iteratively execute Eq. \eqref{k-means} to optimize filters $\textbf{W}_{f_i}$.
\STATE For $k$th sample, obtain $\textbf{z}_i^{(k)}$ according to Eq. \eqref{ConvBLS:z_i^{(k)}} and concatenate the extracted features of all samples as $\textbf{Z}_i$.
\ENDFOR
\STATE Set the feature node groups $\mathbf{Z}=[\textbf{Z}_1,\textbf{Z}_2,...,\textbf{Z}_n]$
\FOR{$j=1$ to $m$}
\STATE Random $\textbf{W}_{e_j}$, $\mathbf{\beta}_{e_j}$
\STATE For $k$th sample, obtain $\textbf{h}_j^{(k)}$ according to Eq. \eqref{ConvBLS:h_j^{(k)}} and concatenate the enhanced features of all samples as $\textbf{H}_j$. 
\ENDFOR
\STATE Set the enhancement node groups $\textbf{H}=[\textbf{H}_1,\textbf{H}_2,...,\textbf{H}_m]$
\STATE For $k$th sample, calculate  $\mathbf{a}^{(k)}$ and concatenate the TSMS features of all samples as $\mathbf{A}^{C_2}$ according to Eq. (\ref{TSMS:a_s1^{(k)}}, \ref{TSMS:a_s2^{(k)}}).
\STATE Calculate $(\mathbf{A}^{C_2})^{+}$ according to Eq. \eqref{moore_pseudo_inverse}.
\WHILE{\textit{The training error threshold is not satisfied}}
\FOR{$j=1$ to $m$}
\STATE Random $\textbf{W}_{e_j}^{C_2+1}$, $\mathbf{\beta}_{e_j}^{C_2+1}$
\STATE Calculate $\mathbf{H}_{j}^{C_2+1}$ according to Eq. \eqref{incre:H_j}.
\ENDFOR
\STATE Obtain $\mathbf{A}^{C_2+1}$ according to Eq. \eqref{incre:A_new}. Calculate $(\mathbf{A}^{C_2+1})^{+}$ and $\mathbf{W}_{out}^{C_2+1}$ according to Eq. (\ref{incre_pi}, \ref{incre:BT}, \ref{incre:W_out}).
\STATE $C_2=C_2+1$
\ENDWHILE
\STATE Set $\mathbf{W}_{out}=\mathbf{W}_{out}^{C_2+1}$
\RETURN $\{\textbf{W}_{f_i}\}_{i=1}^{n}$, $\mathbf{W}_{out}$
\end{algorithmic}
\end{algorithm}
where $\bm{Z}_{0}^{C_1+1}$ is defined as $\bm{X}$. Denote $\bm{Z}^{C_1+1}\equiv[\bm{Z}_{1}^{C_1+1},\bm{Z}_{2}^{C_1+1},...,\bm{Z}_{n}^{C_1+1}]$, the corresponding enhancement nodes are randomly generated as follows:
\begin{multline}
\label{FeaIncre:H_j}
    \bm{H}_{j}^{C_1+1} = \bm{\varphi} (Conv(\bm{H}_{j-1}^{C_1+1}; \{ \bm{W}_{e_j}^{C_1+1}, \bm{\beta}_{e_j}^{C_1+1} \})),\\
    j=1,2,...m
\end{multline}
where $\bm{H}_{0}^{C_1+1}$ is defined as $\bm{Z}^{C_1+1}$. Then, the TSMS features can be expressed as follows:
\begin{small}
\begin{equation}
    \label{FeaIncre:A_new}
    \bm{A}^{C_1+1}= [\bm{A}^{C_1}|\bm{\sigma}(\bm{Z}_{1}^{C_1+1},...,\bm{Z}_{n}^{C_1+1},\bm{H}_{1}^{C_1+1},...,\bm{H}_{m}^{C_1+1})]
\end{equation}
\end{small}
Similarly, we deduce the pseudoinverse of the new matrix as 
\begin{equation}
    \label{FeaIncre:PI}
    (\bm{A}^{C_1+1})^{+}=\begin{bmatrix}
    (\bm{A}^{C_1})^{+}-\bm{DB}\\
    \bm{B}^{\top}
    \end{bmatrix}
\end{equation}
where $\bm{D}=(\bm{A}^{C_1})^{+}\bm{\sigma}(\bm{Z}_{1}^{C_1+1},...,\bm{Z}_{n}^{C_1+1},\bm{H}_{1}^{C_1+1},...,\bm{H}_{m}^{C_1+1})$,
\begin{equation}
    \label{FeaIncre:BT}
    \bm{B}^{\top}=
    \begin{cases}
    (\bm{Q})^{+} &{\text{if}\ \bm{Q} \ne \bm{0}}\\
    (1+\bm{D}^{\top}\bm{D})^{-1}\bm{D}^{\top}(\bm{A}^{C_1})^{+} &{\text{if}\ \bm{Q}=\bm{0}}\\
    \end{cases}
\end{equation}
and $\bm{Q}=\bm{\sigma}(\bm{Z}_{1}^{C_1+1},...,\bm{Z}_{n}^{C_1+1},\bm{H}_{1}^{C_1+1},...,\bm{H}_{m}^{C_1+1})-\bm{A}^{C_1} \bm{D}$,

Again, the new weight matrix is 
\begin{equation}
    \label{FeaIncre:W_out}
    \bm{W}_{out}^{C_1+1}=
    \begin{bmatrix}
    \bm{W}_{out}^{C_1}-\bm{DB}^{\top}\hat{\bm{Y}}\\
    \bm{B}^{\top}\hat{\bm{Y}}\\
    \end{bmatrix}
\end{equation}

The incremental learning algorithm of additional feature nodes is listed in Algorithm \ref{alg:FeatureIncremental}.

Inspired by the great success of deep transfer learning, we also use the pre-trained deep convolutional neural network as the CF layer. Compared with the CF layer trained by the SKM algorithm, using the pre-training model as the CF layer also maintains the training efficiency of BLS and has better performance on more complex tasks.

\section{Experimental results}
In this Section, experimental results are given to verify the proposed ConvBLS. The experiments are conducted on the Ubuntu 20.04 operating system, and the CPU is Intel Xeon Gold 6226R.

\subsection{Dataset}
The experiments are conducted on the following datasets: 1) MNIST, 2) Fashion-MNIST, and 3) NORB.
\subsubsection{MNIST}
The dataset\cite{lecun1998gradient} contains 60 000 training samples and 10 000 test samples, which are evenly distributed over 10 classes of handwritten digital images. Among them, every sample is a gray-scale image with 28$\times$28 pixels.

\subsubsection{Fashion-MNIST}
The dataset\cite{xiao2017fashion} is consistent with the MNIST dataset, except that it has more complex image features. Besides, all samples fall into 10 categories, including 1) T-shirt/top, 2) trouser, 3) pullover, 4) dress, 5) coat, 6) sandal, 7) shirt, 8) sneaker, 9) bag, and 10) ankle boot.

\subsubsection{NORB}
The dataset\cite{lecun2004learning} is a more complicated dataset compared with MNIST and Fashion-MNIST datasets, which is composed of 48 600 images with the size of 2 $\times$ 32 $\times$ 32 pixels. The NORB contains images of 50 different 3-D toy objects labeled by five distinct categories: 1) animals, 2) humans, 3) airplanes, 4) trucks, and 5) cars. Here, 24 300 images of 25 objects are used for training, and the other 24 300 images are used for testing.

\subsection{Performance of ConvBLS}
To investigate the superiority of the proposed model, We perform extensive comparison experiments with four types of methods, including 1) traditional methods, 2) broad topology-based methods, 3) deep topology-based methods, and 4) deep and broad topology-based methods. Unless the relevant papers do not provide valid structural hyperparameters and training details (we marked the training time of these methods with a unique superscript $\dag$), for the sake of reliability and fairness, all the remaining approaches are reproduced on our experimental platform with the same hyperparameters as the original papers, and the corresponding training time is listed. It should be noted that in all experiments, the data augmentation technique is avoided, and we mainly focus on the verification of the effectiveness of the proposed ConvBLS.

The structure parameters of ConvBLS are listed as follows: the number of CF layers is set to 3, and the number of CE layers is set to 1. In addition, the number of pyramid layers in the second phase of the TSMS feature fusion layer is set to 3, and the corresponding number of spatial bins is set to \{3$\times$3,
\begin{algorithm}[H]
\renewcommand{\algorithmicrequire}{\textbf{Input:}}
\renewcommand{\algorithmicensure}{\textbf{Output:}}
\caption{ConvBLS: Increment of $C^a_f$ Additional Feature Nodes in the CF$_1$ Layer}
\label{alg:FeatureIncremental}
\begin{algorithmic}[1]
\REQUIRE training samples $\{\textbf{x}^{(k)}, \hat{\textbf{y}}^{(k)}\}_{k=1}^{N}$.
\ENSURE $\{\textbf{W}_{f_i}\}_{i=1}^{n}$, $\textbf{W}_{out}$.
\FOR{$i=1$ to $n$}
\STATE Group feature maps according to $c_i$. For each feature group, execute patches extraction and preprocessing, iteratively execute Eq. \eqref{k-means} to optimize filters $\textbf{W}_{f_i}$.
\STATE For $k$th sample, obtain $\textbf{z}_i^{(k)}$ according to Eq. \eqref{ConvBLS:z_i^{(k)}} and concatenate the extracted features of all samples as $\textbf{Z}_i$.
\ENDFOR
\STATE Set the feature node groups $\mathbf{Z}=[\textbf{Z}_1,\textbf{Z}_2,...,\textbf{Z}_n]$
\FOR{$j=1$ to $m$}
\STATE Random $\textbf{W}_{e_j}$, $\mathbf{\beta}_{e_j}$
\STATE For $k$th sample, obtain $\textbf{h}_j^{(k)}$ according to Eq. \eqref{ConvBLS:h_j^{(k)}} and concatenate the enhanced features of all samples as $\textbf{H}_j$. 
\ENDFOR
\STATE Set the enhancement node groups $\textbf{H}=[\textbf{H}_1,\textbf{H}_2,...,\textbf{H}_m]$
\STATE For $k$th sample, calculate  $\mathbf{a}^{(k)}$ and concatenate the TSMS features of all samples as $\mathbf{A}^{C_1}$ according to Eq. (\ref{TSMS:a_s1^{(k)}}, \ref{TSMS:a_s2^{(k)}}).
\STATE Calculate $(\mathbf{A}^{C_1})^{+}$ according to Eq. \eqref{moore_pseudo_inverse}.
\WHILE{\textit{The training error threshold is not satisfied}}
\FOR{$i=1$ to $n$}
\STATE Execute step 2 to obtain $\textbf{W}_{f_i}^{C_1+1}$
\STATE Calculate $\bm{Z}_i^{C_1+1}$ according to Eq. \eqref{FeaIncre:Z_i}. 
\ENDFOR
\STATE Set the additional feature node groups $\bm{Z}^{C_1+1}=[\bm{Z}_{1}^{C_1+1},\bm{Z}_{2}^{C_1+1},...,\bm{Z}_{n}^{C_1+1}]$
\FOR{$j=1$ to $m$}
\STATE Random $\textbf{W}_{e_j}^{C_1+1}$, $\mathbf{\beta}_{e_j}^{C_1+1}$
\STATE \vspace{0.5pt}Calculate $\mathbf{H}_{j}^{C_1+1}$ according to Eq. \eqref{FeaIncre:H_j}.
\ENDFOR
\STATE Obtain $\bm{A}^{C_1+1}$ according to Eq. \eqref{FeaIncre:A_new}. Calculate $(\mathbf{A}^{C_1+1})^{+}$ and $\mathbf{W}_{out}^{C_1+1}$ according to Eq. (\ref{FeaIncre:PI}, \ref{FeaIncre:BT}, \ref{FeaIncre:W_out}).
\STATE $C_1=C_1+1$
\ENDWHILE
\STATE Set $\mathbf{W}_{out}=\mathbf{W}_{out}^{C_1+1}$, $\{\mathbf{W}_{f_i}\}_{i=1}^{n}=\{[\mathbf{W}_{f_i}|\mathbf{W}_{f_i}^{C_1+1} ]\}_{i=1}^{n}$
\RETURN $\{\textbf{W}_{f_i}\}_{i=1}^{n}$, $\mathbf{W}_{out}$
\end{algorithmic}
\end{algorithm}
2$\times$2, 1$\times$1\}. After that, the remaining structural parameters depend on the complexity of the specific task. As for the training-related parameters, the number of patches extracted from the output of the previous CF layer is set to 400 000. The regularization coefficient $\lambda$ for the output layer is chosen from the set $\{10^{-5}, 10^{-4}, ..., 10^{5}\}$.

\subsubsection{Experimental Results on MNIST Dataset}
Our results are shown in Table \ref{tab:mnist}. The experimental results of the comparison methods, including SAE, DBN, and MLELM, are cited from \cite{tang2015extreme}, while that of Stacked BLS are cited from \cite{liu2020stacked}. For our ConvBLS, we set the initial number of output channels in the CF$_1$ layer as $32$ and the corresponding expansion ratio as $2$.

We can observe that ConvBLS has the highest test accuracy of $99.280\%$, even with the extremely short training time. Specifically, our method have a speedup of more than $200$ times compared to the most time-consuming method, which is attributed to the TSMS feature extracted by the effective model architecture and the efficient training algorithm.
\begin{table*}[!t]
\caption{Experimental Results on MNIST Dataset\label{tab:mnist}}
\centering
\begin{tabular}{c c c c c}
\hline\hline
Method & Test Accuracy (\%) & Training Times (s) & Speedup Times & Topology\\
\hline
SAE \cite{tang2015extreme} & 98.60 & 36448.40$^\dag$ & 1.46014 & - \\
DBN \cite{tang2015extreme} & 98.87 & 53219.77$^\dag$ & 1 & - \\
MLELM \cite{tang2015extreme} & 99.04 & 475.83$^\dag$ & 111.8462 & - \\
\hline
BLS \cite{chen2017broad} & 98.740 & 47.3725 & 1123.432 & broad \\
CFEBLS \cite{chen2018universal} & 98.83 & 24.1333 & 2205.242 & broad \\
R-BLS \cite{zhang2018rich} & 98.95 & - & - & broad \\
CFBLS-pyramid \cite{zhang2020analysis} & 98.65 ± 0.14 & 64.4368 & 825.922 & broad \\
Stacked BLS \cite{liu2020stacked} & 99.120 & 30.1916$^\dag$ & 1762.734 & broad \\
\hline
MLP \cite{bishop2006pattern} & 97.39 & 633.8427 & 83.96369 & deep \\
LeNet-5 \cite{lecun1989handwritten} & 95.63 & 732.8154 & 72.62371 & deep \\
ResNet34 \cite{he2016deep} & 98.960 & 20469.234 & 2.599988 & deep \\
\hline
CNNBLS \cite{yang2018cnn} & 96.940 & 377.4365 & 141.0032 & deep + broad \\
CNN + BLS \cite{li2019cnn} & 99.230 & 790.492 & 67.32487 & deep + broad \\
Ours & 99.280 & 228.8355 & 232.5678 & deep + broad \\
\hline\hline
\end{tabular}
\end{table*}

\begin{table*}[!t]
\caption{Experimental Results on Fashion-MNIST Dataset\label{tab:fashion_mnist}}
\centering
\begin{tabular}{c c c c c}
\hline\hline
Method & Test Accuracy (\%) & Training Times (s) & Speedup Times & Topology\\
\hline
KNN \cite{zhang2020analysis} & 84.70 ± 0.00 & 4927$^\dag$ & 6.759972 & - \\
RF \cite{liu2020stacked} & 87.3 & - & - & - \\
Xgboost \cite{liu2020stacked} & 89.82 & - & - & - \\
Dyra-Net \cite{liu2020stacked} & 90.6 & - & - & - \\
\hline
BLS \cite{chen2017broad} & 91.39 & 46.6083 & 714.6019 & broad \\
CFEBLS \cite{chen2018universal} & 87.130 & 24.5927 & 1354.32 & broad \\
R-BLS \cite{zhang2018rich} & 87.48 & - & - & broad \\
CFBLS-pyramid \cite{zhang2020analysis} & 89.88 ± 0.15 & 66.4128 & 501.5054 & broad \\
Stacked BLS \cite{liu2020stacked}
& 91.53 & - & - & broad \\
\hline
AlexNet \cite{krizhevsky2012imagenet} & 87.1 & 1016.595 & 32.76268 & deep \\
VGG16 \cite{simonyan2014very} & 90.28 & 6400.780 & 5.203488 & deep \\
GoogLeNet \cite{szegedy2015going} & 91.75 & 7792.025 & 4.274419 & deep \\
DenseNet \cite{huang2017densely} & 90.75 & 33306.380 & 1 & deep \\
\hline
CNNBLS \cite{yang2018cnn} & 84.210 & 576.8780 & 57.73557 & deep + broad \\
CNN + BLS \cite{li2019cnn} & 91.170 & 826.8948 & 40.27886 & deep + broad \\
Ours & 92.430 & 332.7186 & 100.1038 & deep + broad \\
\hline\hline
\end{tabular}
\end{table*}

\begin{table*}[!t]
\caption{Experimental Results on NORB Dataset\label{tab:norb}}
\centering
\begin{tabular}{c c c c c}
\hline\hline
Method & Test Accuracy (\%) & Training Times (s) & Speedup Times & Topology\\
\hline
K-means (Triangle) + SVM \cite{coates2011analysis} & 97.0 & 433.7307 & 97.83569 & - \\
\hline
BLS \cite{chen2017broad} & 89.27 & 11.9946 & 3537.787 & broad \\
CFEBLS \cite{chen2018universal} & 90.02 & 20.8354 & 2036.646 & broad \\
CEBLS-dense \cite{zhang2020analysis} & 88.40 ± 0.29 & 37.1291 & 1142.886 & broad \\
K-means-BLS \cite{liu2017broad} & 95.971 & 309.9045 & 136.9272 & broad \\
Stacked BLS \cite{liu2020stacked} & 91.90 & 5.1718$^\dag$ & 8204.946 & broad \\
\hline
MLP \cite{bishop2006pattern} & 85.325 & 299.4026 & 141.73 & deep \\
LeNet-5 (ReLU) \cite{lecun1989handwritten} & 87.453 & 212.2769 & 199.9009 & deep \\
AlexNet-small \cite{krizhevsky2012imagenet} & 91.218 & 594.974 & 71.32134 & deep \\
AlexNet-base \cite{krizhevsky2012imagenet} & 91.930 & 1107.2195 & 38.32514 & deep \\
AlexNet-large \cite{krizhevsky2012imagenet} & 93.049 & 3497.686 & 12.13212 & deep \\
VGG13 \cite{simonyan2014very} & 96.486 & 4380.396 & 9.68733 & deep \\
ResNet18 \cite{he2016deep} & 94.646 & 12658.889 & 3.352138 & deep \\
ResNet34 \cite{he2016deep} & 93.988 & 18975.880 & 2.236225 & deep \\
ResNet50 \cite{he2016deep} & 95.045 & 42434.341 & 1 & deep \\
\hline
CNNBLS \cite{yang2018cnn} & 90.066 & 218.7023 & 194.0279 & deep + broad \\
CNN+BLS \cite{li2019cnn} & 91.016 & 279.224 & 151.9724 & deep + broad \\
Ours & 97.193 & 194.4663 & 218.2092 & deep + broad \\
\hline\hline
\end{tabular}
\end{table*}

\subsubsection{Experimental Results on Fashion-MNIST Dataset}
Table \ref{tab:fashion_mnist} presents the results. To make the experimental conclusions more reliable, the test accuracy of several comparison methods, including AlexNet, VGG16, GoogLeNet, and DenseNet, are cited from \cite{meshkini2019analysis} and \cite{duan2019image}. Conversely, the training time of these models are obtained by re-running them on our experimental platform with the same training details as the original papers for a fair comparison. For CNNBLS and CNN+BLS, the model structures and training details we adopted are the same as that on the MNIST dataset. At last, given the more challenging data set, the number of initial output channels and expansion ratio was set to $64$ and $1.5$, respectively.

As shown in Table \ref{tab:fashion_mnist}, similar to the MNIST dataset case, our method achieve the state-of-the-art performance among the existing approaches with a superfast speed in computation. Thus, the proposed ConvBLS model is very appealing. In particular, our approach surpasses many classical DNNs in both time and accuracy.
\begin{table*}[!t]
\caption{Experimental Results of Incremental Learning on MNIST, Fashion-MNIST and NORB datasets\label{tab:increment}}
\centering
\begin{tabular}{c c c c c c c}
\hline\hline
Datasets & \makecell[c]{Incremental \\Algorithm} & \makecell[c]{Number of Initial \\Channels in \\the CF$_1$ Layer} & \makecell[c]{Number of Initial \\Channels in \\the CE$_1$ Layer} & Test Accuracy (\%) & \makecell[c]{Additional Training \\Time (s)} & \makecell[c]{Accumulative Training \\Time (s)} \\
\hline
\multirow{10}{*}{MNIST} & \multirow{5}{*}{\makecell[c]{Feature \\Nodes}} & 16 & 224 & 98.850 & 150.1506 & 150.1506 \\
 & & 16 $\rightarrow$ 20 & 224 $\rightarrow$ 280 & 98.930 & 68.9228 & 219.0734 \\
 & & 20 $\rightarrow$ 24 & 280 $\rightarrow$ 336 & 98.950 & 69.5863 & 288.6597 \\
 & & 24 $\rightarrow$ 28 & 336 $\rightarrow$ 392 & 98.990 & 70.2971 & 358.9568 \\
 & & 28 $\rightarrow$ 32 & 392 $\rightarrow$ 448 & 99.050 & 70.7071 & 429.6578 \\
\cline{2-7}
 & \multirow{5}{*}{\makecell[c]{Enhancement \\Nodes}} & 16 & 112 & 98.670 & 114.5602 & 114.5602 \\
 & & 16 & 112 $\rightarrow$ 140 & 98.720 & 1.7916 & 116.3518 \\
 & & 16 & 140 $\rightarrow$ 168 & 98.760 & 1.9123 & 118.2641 \\
 & & 16 & 168 $\rightarrow$ 196 & 98.770 & 1.9750 & 120.2391 \\
 & & 16 & 196 $\rightarrow$ 224 & 98.790 & 2.1035 & 122.3426 \\
\hline
\multirow{10}{*}{Fashion-MNIST} & \multirow{5}{*}{\makecell[c]{Feature \\Nodes}} & 32 & 228 & 91.680 & 216.3523 & 216.3523 \\
 & & 32 $\rightarrow$ 40 & 228 $\rightarrow$ 285 & 91.900 & 88.5769 & 304.9292 \\
 & & 40 $\rightarrow$ 48 & 285 $\rightarrow$ 342 & 92.410 & 92.5383 & 397.4675 \\
 & & 48 $\rightarrow$ 56 & 342 $\rightarrow$ 399 & 92.670 & 95.6273 & 493.0948 \\
 & & 56 $\rightarrow$ 64 & 399 $\rightarrow$ 456 & 92.750 & 92.1845 & 585.2793\\
\cline{2-7}
 & \multirow{5}{*}{\makecell[c]{Enhancement \\Nodes}} & 32 & 114 & 91.390 & 166.2956 & 166.2956 \\
 & & 32 & 114 $\rightarrow$ 142 & 91.570 & 2.0621 & 168.3577 \\
 & & 32 & 142 $\rightarrow$ 170 & 91.760 & 2.2059 & 170.5636 \\
 & & 32 & 170 $\rightarrow$ 198 & 91.810 & 2.2369 & 172.8005 \\
 & & 32 & 198 $\rightarrow$ 226 & 91.940 & 2.6206 & 175.4211 \\
\hline
\multirow{10}{*}{NORB} & \multirow{5}{*}{\makecell[c]{Feature \\Nodes}} & 32 & 228 & 95.564 & 149.4078 & 149.4078 \\
 & & 32 $\rightarrow$ 40 & 228 $\rightarrow$ 285 & 95.704 & 68.5435 & 217.9513 \\
 & & 40 $\rightarrow$ 48 & 285 $\rightarrow$ 342 & 95.852 & 71.7318 & 289.6831 \\
 & & 48 $\rightarrow$ 56 & 342 $\rightarrow$ 399 & 96.296 & 70.7071 & 360.3902 \\
 & & 56 $\rightarrow$ 64 & 399 $\rightarrow$ 456 & 96.811 & 70.2468 & 430.6370 \\
 \cline{2-7}
 & \multirow{5}{*}{\makecell[c]{Enhancement \\Nodes}} &  32 & 114 & 95.547 & 120.3834 & 120.3834 \\
 & & 32 & 114 $\rightarrow$ 142 & 95.560 & 1.0154 & 121.3988 \\
 & & 32 & 142 $\rightarrow$ 170 & 95.593 & 1.0771 & 122.4759 \\
 & & 32 & 170 $\rightarrow$ 198 & 95.675 & 1.0751 & 123.5510 \\
 & & 32 & 198 $\rightarrow$ 226 & 95.712 & 1.1784 & 124.7294 \\
\hline\hline
\end{tabular}
\end{table*}

\subsubsection{Experimental Results on NORB Dataset}
The experimental results are reported in Table \ref{tab:norb}. In the first methods, the k-means and triangle activation are used for single-layer feature extraction, and a support vector machine is used for classification, which generally requires a large number of features.
For reference, K-means-BLS, one of the most similar works to ours, that use k-means to extract features and use a complete BLS to classify features, are reproduced on the NORB dataset, and the relevant hyper-parameters are as follows: the number of features in K-means feature extraction is set to $1600$, and BLS is constructed by total $100 \times 10$ feature nodes and $1 \times 9000$ enhancement nodes. Moreover, to our knowledge, no work attempts to perform classification tasks on NORB datasets using typical CNNs. Therefore, we instantiate some common CNNs as deep topology-based comparison methods. In the first convolutional layer, AlexNet-small, AlexNet-base, and AlexNet-large have $16$, $32$, and $64$ output channels, respectively, and the number of output channels of the remaining layers vary accordingly. For the comparison methods trained by the BP algorithm, similar to \cite{tang2015extreme} and \cite{chen2017broad}, the training procedure is set as $100$ epochs, and the remaining training hyperparameters are finetuned to make the models converge. Lastly, the ConvBLS structure keeps same as that on Fashion-MNIST dataset.
 
The results also indicate that ConvBLS outperforms all of the comparison methods.
In particular, our method significantly outperforms the broad topology-based and deep topology-based comparison methods in terms of test accuracy with a very short training time, which also validates our conjecture that convolution is very effective in processing image data, yet the arduous BP-based training for convolution filters is not necessary on relatively small-scale image datasets.

\subsection{Incremental Learning Experiments}
To illustrate the incremental learning ability of ConvBLS, incremental learning experiments are conducted, and the results are reported in Table \ref{tab:increment}. In this part, we perform two types of incremental experiments: 1) feature node increments and 2) enhancement node increments. All the above experiments are performed on the MNIST, Fashion-MNIST, and NORB datasets. 

Since the experimental setup is similar on the three datasets, next, we present the incremental experiments on the MNIST dataset as an example. For the experiments of increments of feature nodes, the initial number of feature nodes and enhancement nodes in the CF$_1$ layer and the CE$_1$ layer is $16$ and $224$, respectively. In the incremental process, $4$ feature nodes and $56$ enhancement nodes are added each time. For the experiments of increments of enhancement nodes, it initially has $16$ feature nodes and $112$ enhancement nodes in the CF$_1$ layer and CE$_1$ layer, and each incremental step inserts $28$ enhancement nodes.

In Table \ref{tab:increment}, the first conclusion is that the two incremental learning algorithms are effective. As the number of feature nodes or enhancement nodes increases, the test accuracy gradually improves at the cost of acceptable additional training
\begin{figure*}[!t]
    \centering
    \subfloat[]{
        \includegraphics[width=2.1in]{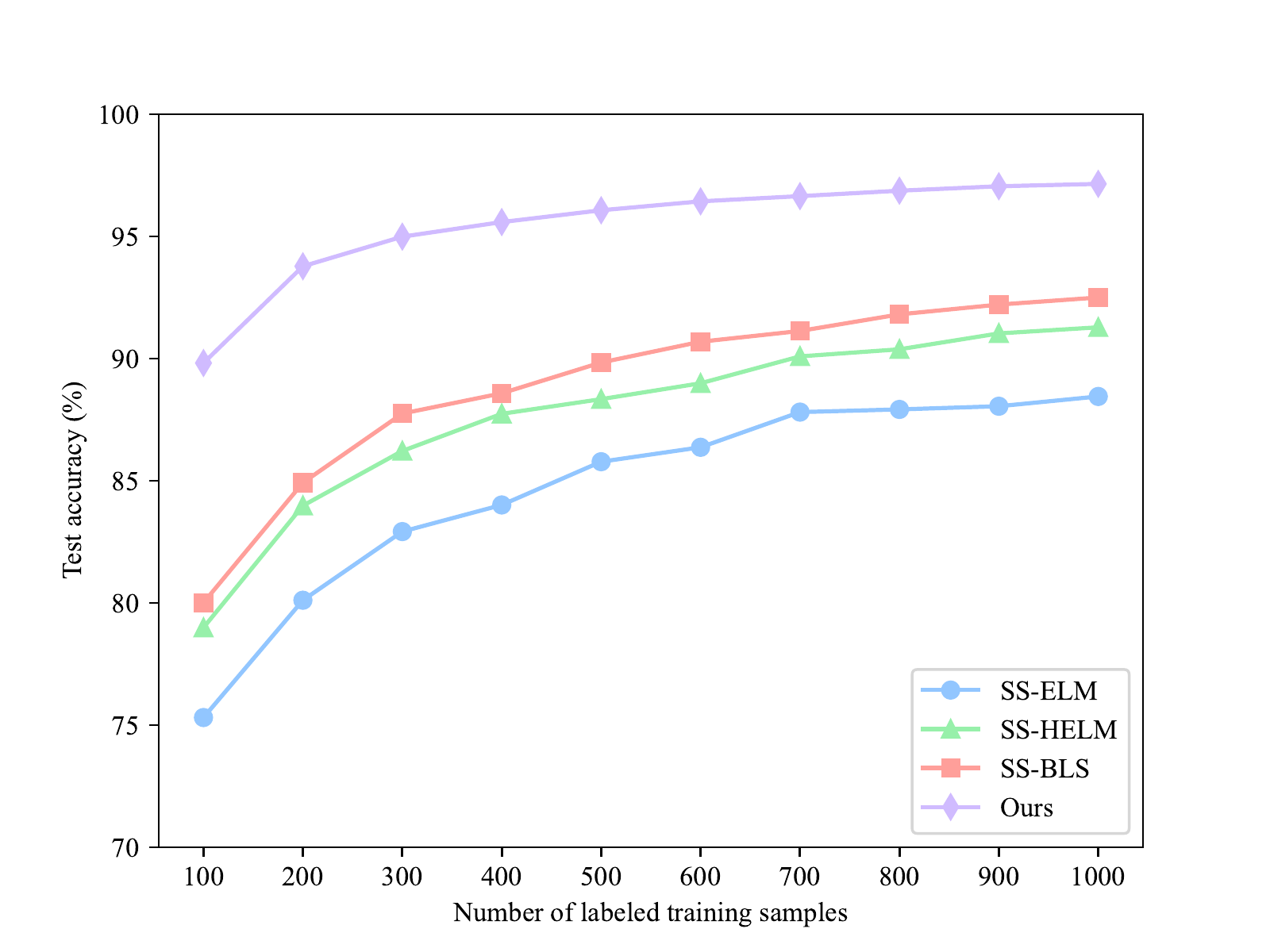}
        \label{fig:semi-supervised:a}
    }
    \hfil
    \subfloat[]{
        \includegraphics[width=2.1in]{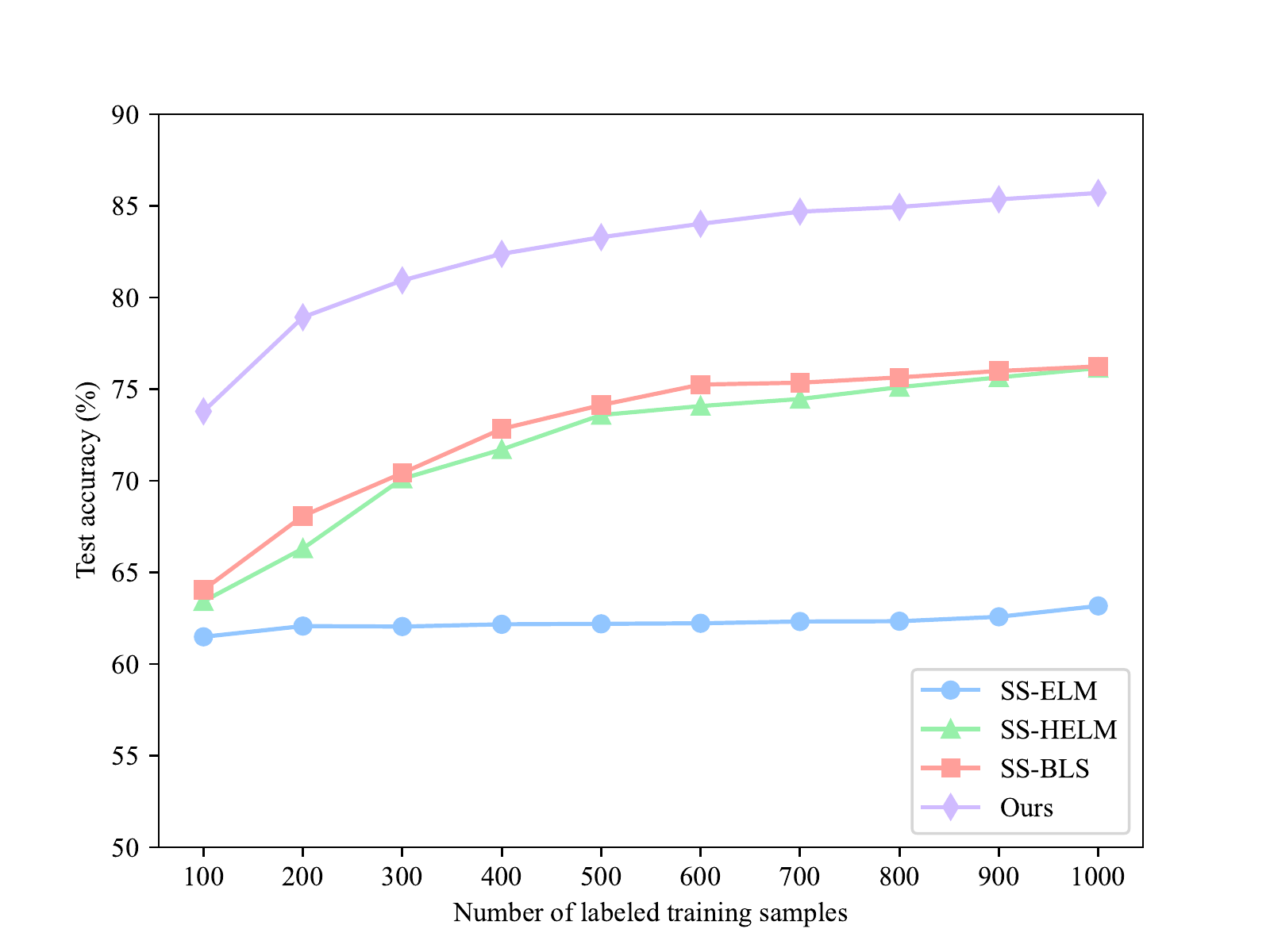}
        \label{fig:semi-supervised:b}
    }
    \hfil
    \subfloat[]{
        \includegraphics[width=2.1in]{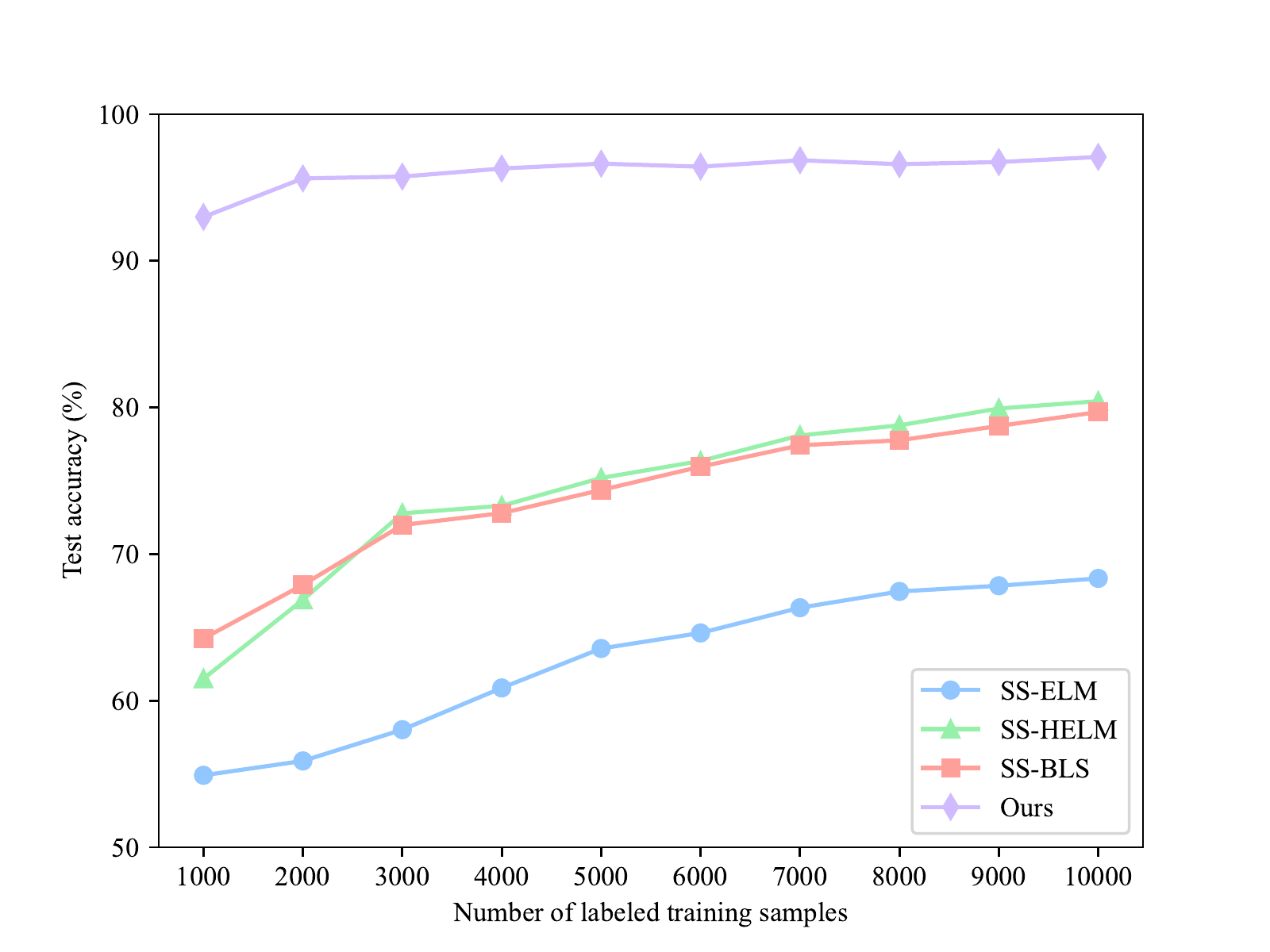}
        \label{fig:semi-supervised:c}
    }
    \newline
    \subfloat[]{
        \includegraphics[width=2.1in]{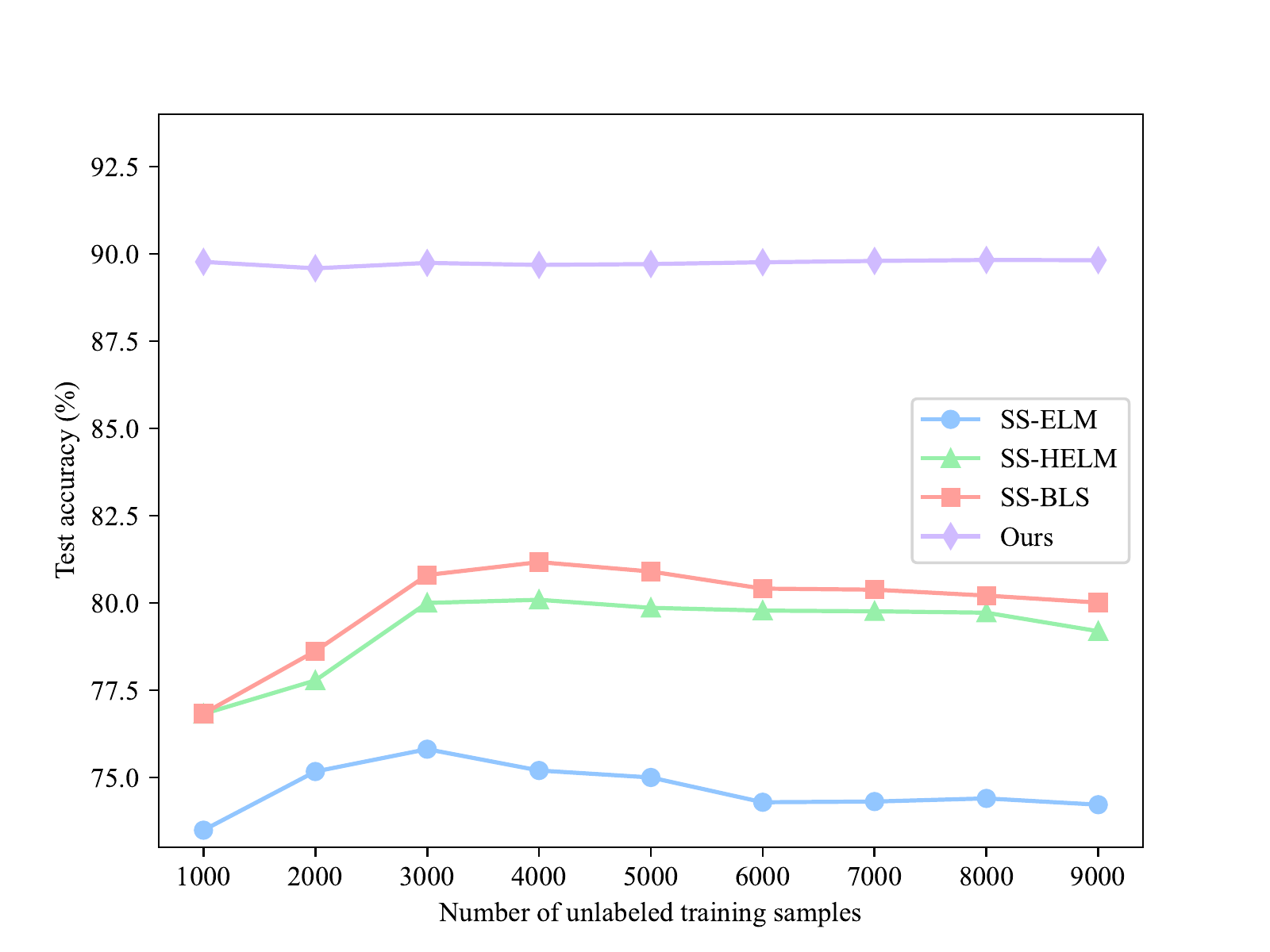}
        \label{fig:semi-supervised:d}
    }
    \hfil
    \subfloat[]{
        \includegraphics[width=2.1in]{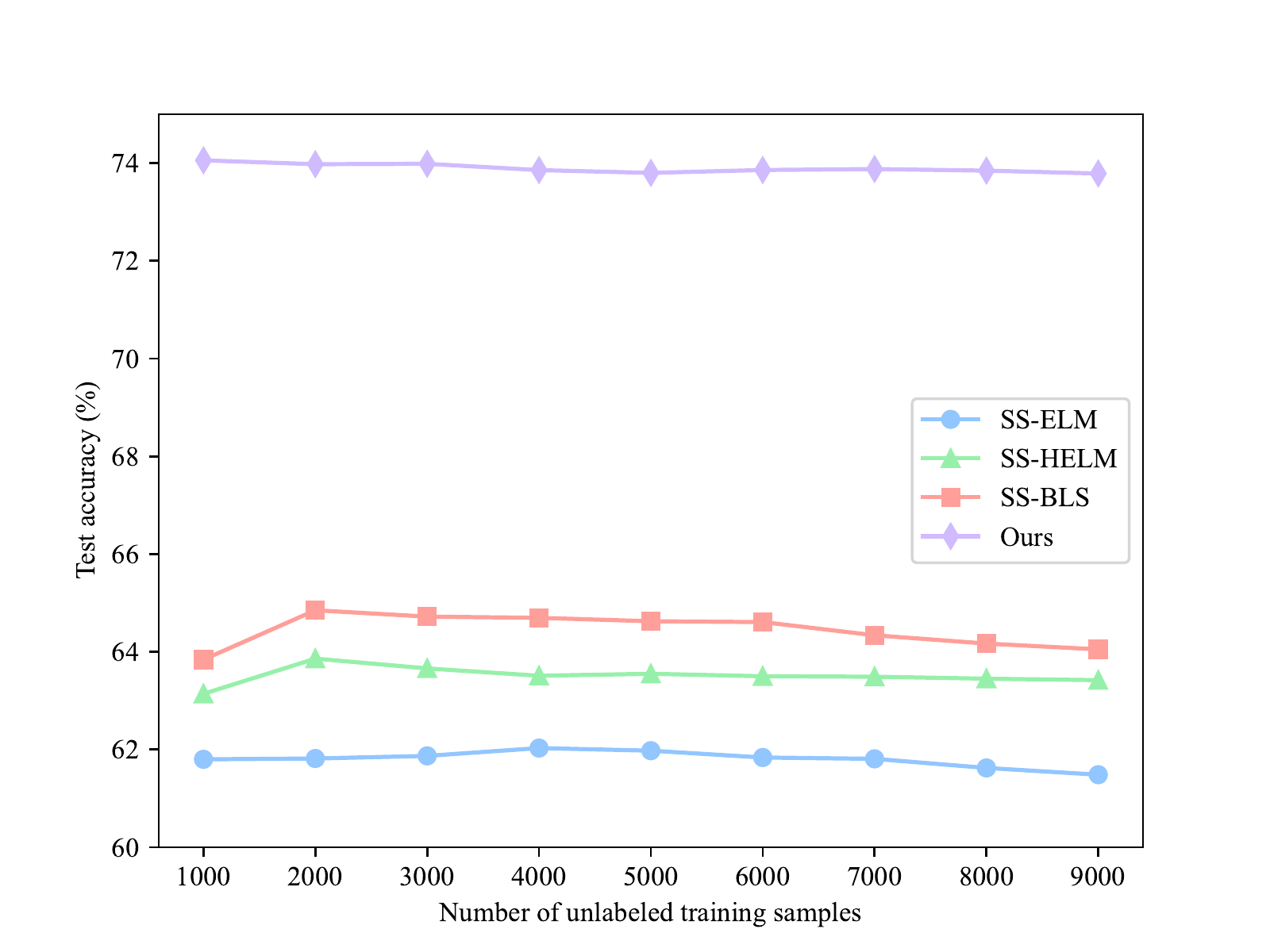}
        \label{fig:semi-supervised:e}
    }
    \hfil
    \subfloat[]{
        \includegraphics[width=2.1in]{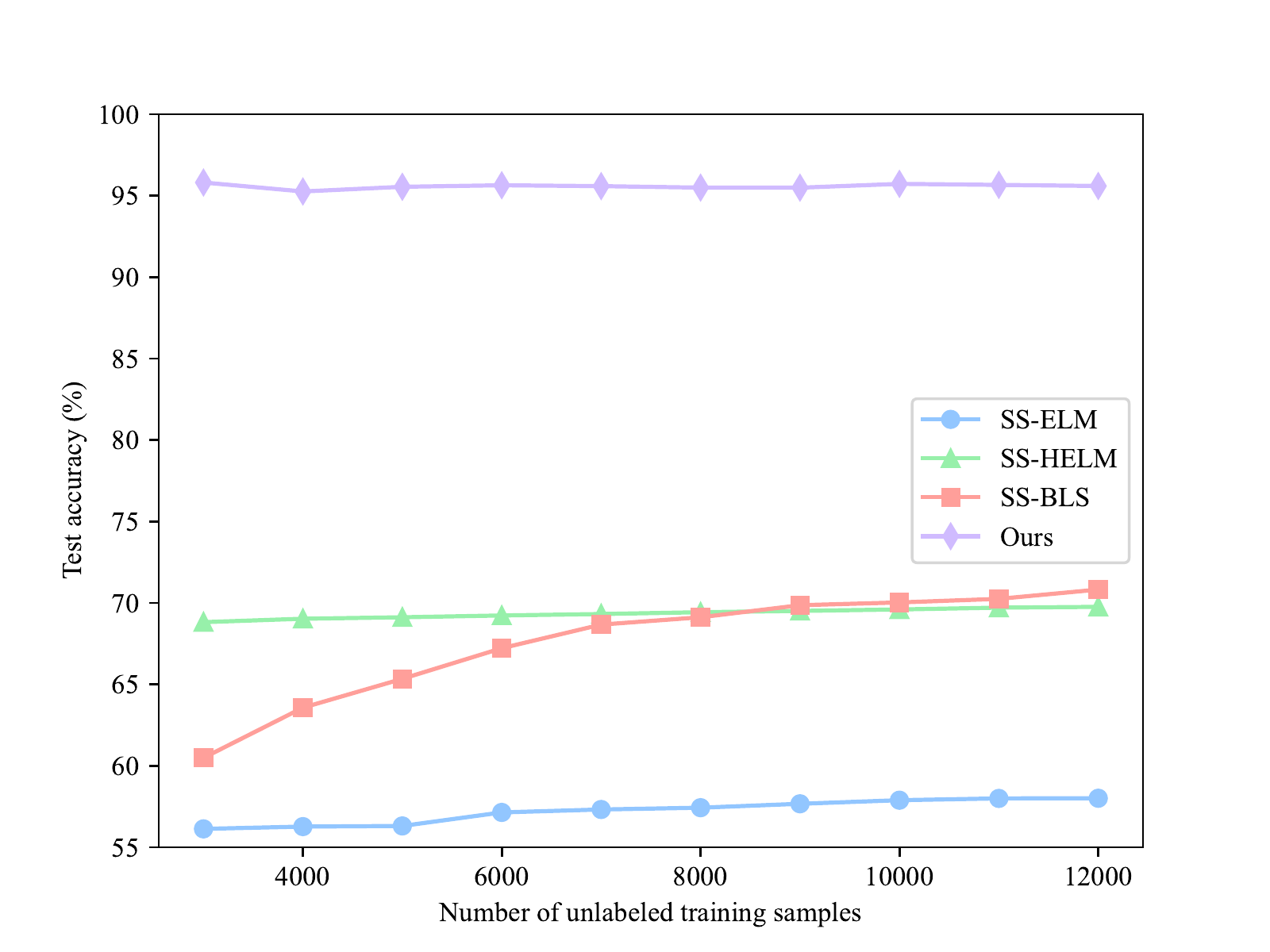}
        \label{fig:semi-supervised:f}
    }
    \newline
    \caption{Classification accuracy of semi-supervised tasks on MNIST, Fashion-MNIST and NORB datasets. (a) The test accuracy with increasing of labeled training samples on MNIST dataset. (b) The test accuracy with increasing of labeled training samples on Fashion-MNIST dataset. (c) The test accuracy with increasing of labeled training samples on NORB dataset. (d) The test accuracy with increasing of unlabeled training samples on MNIST dataset. (e) The test accuracy with increasing of unlabeled training samples on Fashion-MNIST dataset. (f) The test accuracy with increasing of unlabeled training samples on NORB dataset.}
    \label{fig:semi-supervised}
\end{figure*}
time. Subsequently, comparing the two incremental learning algorithms, we can see that incremental learning for enhancement nodes is more efficient because the weights of CE layers do not need to be trained. Therefore, we can first increment the CE layer when we consider expanding our model.

\subsection{Semi-supervised Learning Experiments}
To verify the superiority of our ConvBLS in semi-supervised classification, we perform extensive experiments following the experimental design of SS-BLS\cite{zhao2020semi}, which is the most typical method to modify BLS for semi-supervised scenarios. First, on the MNIST dataset, to analyze the performance of ConvBLS as the number of labeled training samples increases, the initialized number of labeled training samples is 100, the number of unlabeled training samples is 9000, and the number of labeled test samples is 60 000. Then, the number of labeled training samples is increased by 100 each time until 1000. To analyze the performance as the number of unlabeled training samples increases, the number of labeled training samples is 100, the number of initial unlabeled training samples is 1000, and the number of labeled test samples is 60 000. And then, the number of unlabeled training samples is increased by 1000 each time until 9000. Second, on the Fashion-MNIST dataset, the experimental design is the same as that on the MNIST dataset. Lastly, two sets of experiments are conducted on the NORB dataset. For the first one, the number of initial labeled training samples is 1000, the number of unlabeled training samples is 14 300, and the number of labeled test samples is 24 300. Subsequently, the number of labeled training samples is increased by 1000 each time until 10 000. For the second one, the number of labeled training samples is 2500, the number of initial unlabeled training samples is 3000, and the number of labeled test samples is 24 300. After that, the number of unlabeled training samples is increased by 1000 each time until 12 000. 

The results of all comparison methods on MNIST and NORB datasets are cited from \cite{zhao2020semi}, and that on the Fashion-MNIST dataset is obtained by reproducing them under the above setting. Similarly, the structural and regularization parameters are obtained by grid search. For our ConvBLS, all parameters are identical to that in Section IV.B. To eliminate the random factor, we run each experiment 10 times independently, and their means are selected as the final performance.

As depicted in Fig. \ref{fig:semi-supervised}, the testing accuracy of ConvBLS is significantly higher than that of comparison methods in all settings. Moreover, the test accuracy can gradually improve with the increase of the number of labeled training samples, which does not hold in Fig. \ref{fig:semi-supervised}(d)-(f) yet. Thus, it may indicate that despite the ingenious semi-supervised framework developed by the comparison methods, almost all of them still fail to take full advantage of the information from the unlabeled data. Consequently, we argue that richer and more 
comprehensive semantic features are as important as elaborate semi-supervised framework designs, which typically use complex graph theory and manifold regularization, rather than sample unsupervised feature learning.

\subsection{Ablation Study}
To investigate the necessities and effectiveness of each component of ConvBLS, ablation experiments fall into three parts. The first one verifies the effectiveness of the unsupervised training algorithm for CF layers. Second, experiments are designed to demonstrate the effectiveness of TSMS feature fusion. At last, we progressively construct our model to achieve increasing performance.

\begin{figure}[!t]
\centering
\includegraphics[width=2.8in]{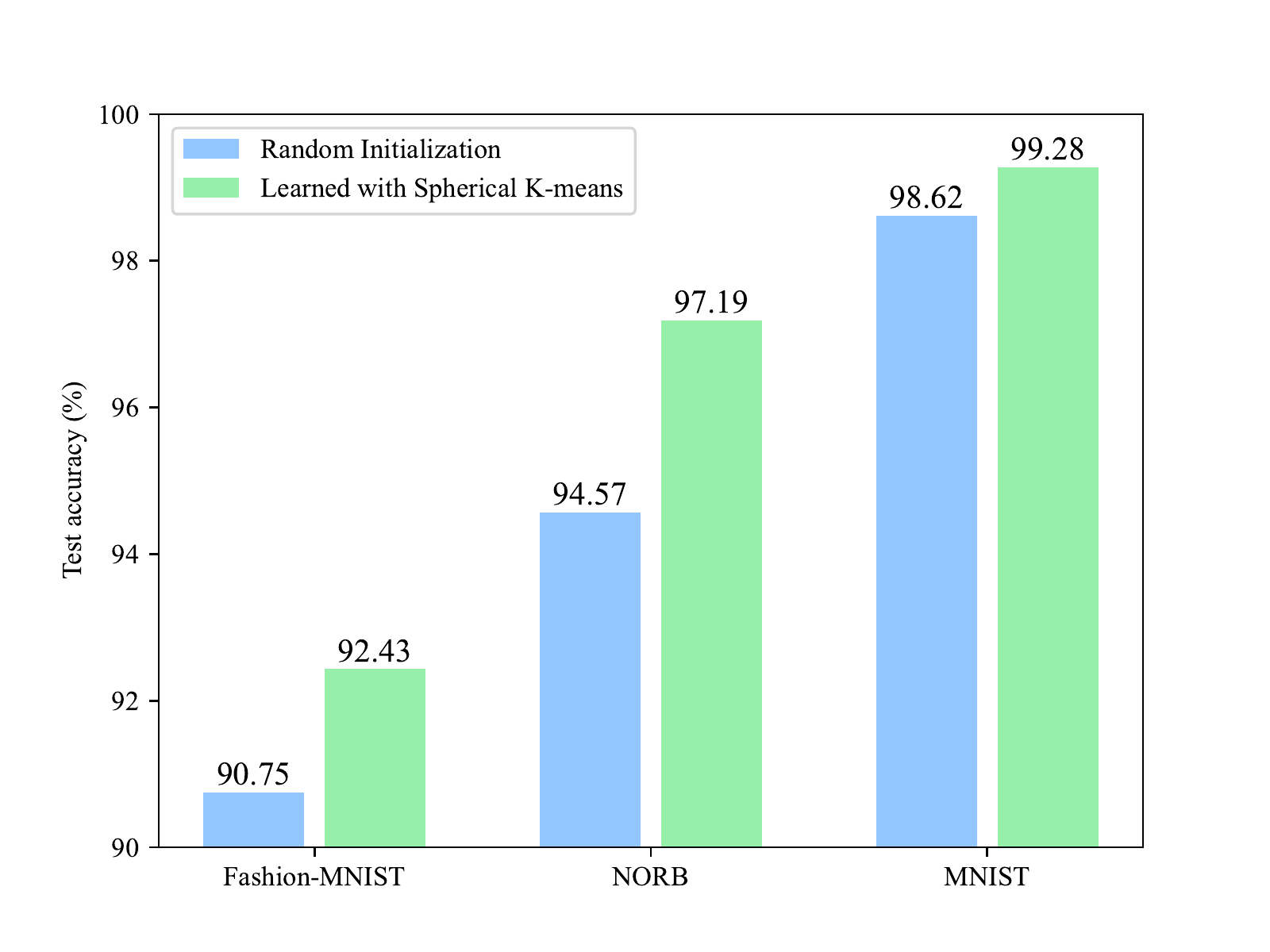}
\caption{Comparison results of the ablation experiment on spherical K-means.}
\label{fig:ablation:kmeans}
\end{figure}

\begin{figure}[!t]
    \centering
    \subfloat[]{
        \includegraphics[width=1.3in]{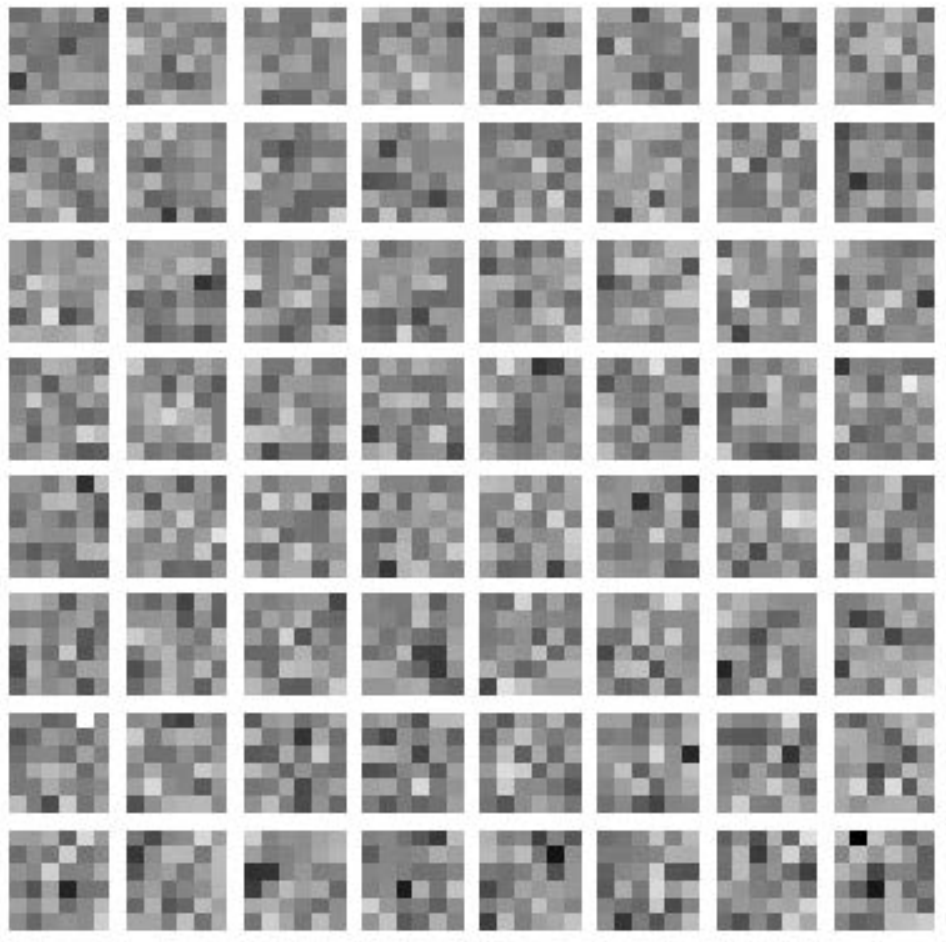}
        \label{fig_first_label}
    }
    \hfil
    \subfloat[]{
        \includegraphics[width=1.3in]{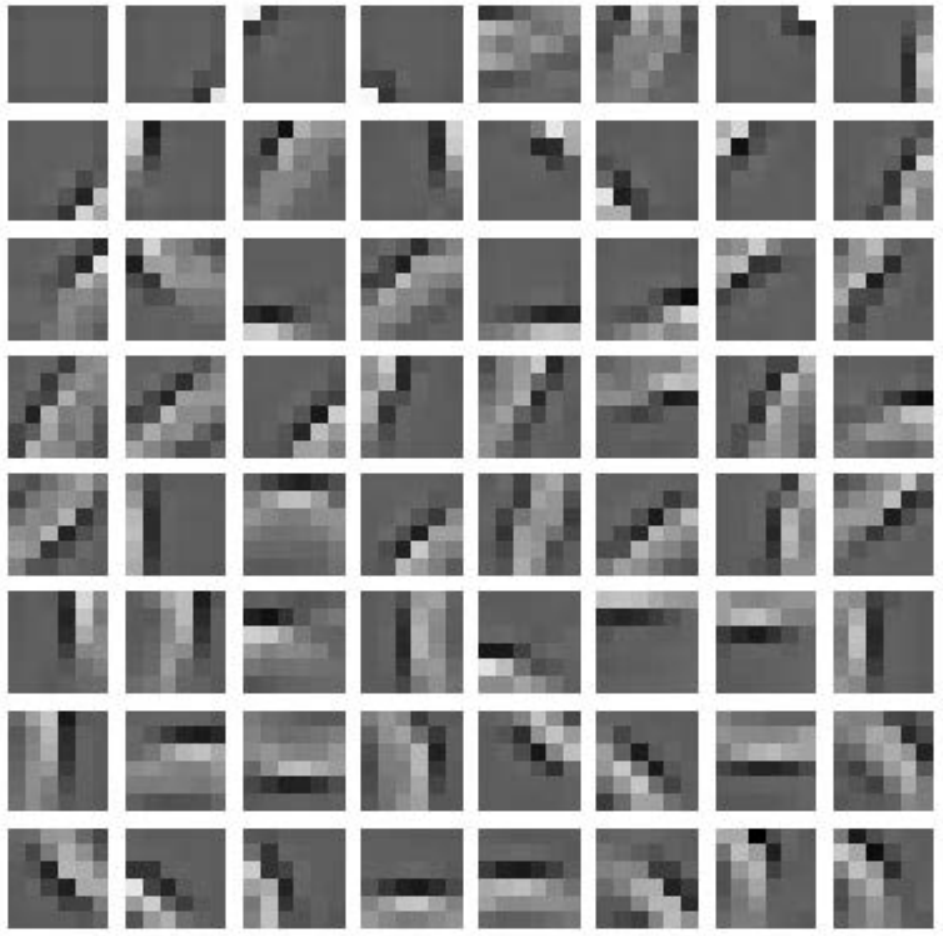}
        \label{fig_first_label}
    }
    \caption{Filters from the CF$_1$ layer on the FashionMNIST dataset. (a) Random initialization. (b) Learned with spherical K-means.}
    \label{fig:filters}
\end{figure}

In the first part, the comparison method is de facto identical to our ConvBLS except that the filter parameters of CF layers are randomly generated. Additionally, hyperparameters of both the comparison method and our ConvBLS are consistent with that in Section IV.B. As shown in Fig. \ref{fig:ablation:kmeans}, our ConvBLS outperforms the comparison method on all of the three datasets, which indicate that the unsupervised learning algorithm we designed for CF layers is effective.
To investigate the reason why the filters trained using the SKM significantly outperform the random filters, we visualize the filters of the CF$_1$ layer for both our method and comparison method on the FashionMNIST dataset as an example. As illustrated in Fig. \ref{fig:filters}, compared to the randomly initialized filters, the filters obtained using the SKM are more like, to some extent, edge detectors and corner detectors.
This distinct selectivity for local features is crucial, especially in the image feature extraction procedure. With these merits, our methods can achieve better accuracies than comparison methods.

\begin{figure}[!t]
\centering
\includegraphics[width=2.8in]{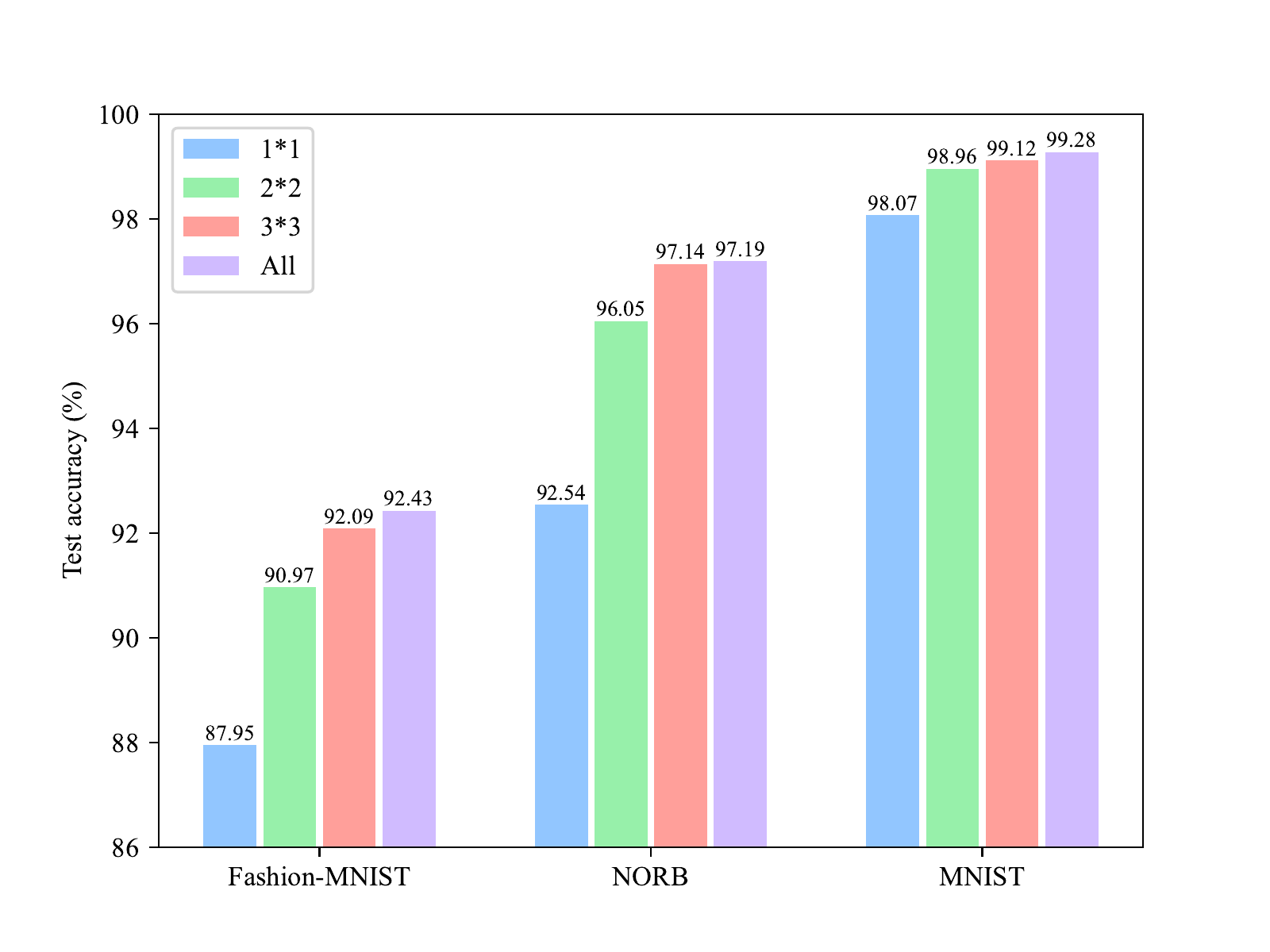}
\caption{Comparison results of the ablation experiment on the TSMS feature fusion.}
\label{fig:ablation:spp}
\end{figure}

Secondly, we conduct a set of ablation experiments on the TSMS feature fusion. For each dataset, four models with 1*1 bins, 2*2 bins, 3*3 bins, and all three pyramid features are trained and evaluated. Fig. \ref{fig:ablation:spp} shows that pyramid layers with more spatial bins have higher accuracy, and the spatial pyramids with all the above pyramid layers have the highest accuracy. 
Note that the method with a three-level pyramid has TSMS features, while the comparison methods do not. Thus, we can safely conclude that the TSMS features are more robust and effective for image classification tasks.

Lastly, we add modules proposed in this research one by one and construct progressively three individual ConvBLS models. The results are presented in Table \ref{tab:ablation}. The hyperparameters are identical to that in Section III.B, except specifically mentioned in this table. We can see that the most complete model has the best accuracy, and each module is beneficial for performance improvement.

\begin{table*}[!t]
\caption{Ablation Study of the Proposed Method on MNIST, Fashion-MNIST, and NORB dataset. \label{tab:ablation}}
\centering
\begin{tabular}{c c c c c | c c c}
\hline\hline
1 CF Layer & 2 CF Layers & 3 CF Layers & \makecell[c]{3 CF Layers with \\CE Layers} & \makecell[c]{3 CF Layers with \\CE Layers and TSMS} & MNIST & Fashion-MNIST & NORB\\
\hline
\checkmark & & & & & 97.81\% & 88.50\% & 93.580\% \\
\checkmark & \checkmark & & & & 98.45\% & 89.62\% & 93.963\% \\
\checkmark & \checkmark & \checkmark & & & 98.56\% & 89.65\% & 95.132\% \\
\checkmark & \checkmark & \checkmark & \checkmark & & 98.96\% & 90.97\% & 95.132\% \\
\checkmark & \checkmark & \checkmark & \checkmark & \checkmark & 99.28\% & 92.43\% & 97.193\% \\
\hline\hline
\end{tabular}
\end{table*}

\subsection{The Hyper-parameter Sensitivity Analysis of the ConvBLS}
We analyze the sensitivity of the hyper-parameters, including kernel size, stride, activation function, and the number of output channels in ConvBLS by testing how the hyper-parameters influence performance. In the following sections, except for the hyper-parameters we are analyzing, the others are identical to that in the Section IV.B.
As an example, all the experiments below are conducted on the NORB dataset.

\subsubsection{The effect of output channels}
\begin{figure}[!t]
\centering
\includegraphics[width=2.78in]{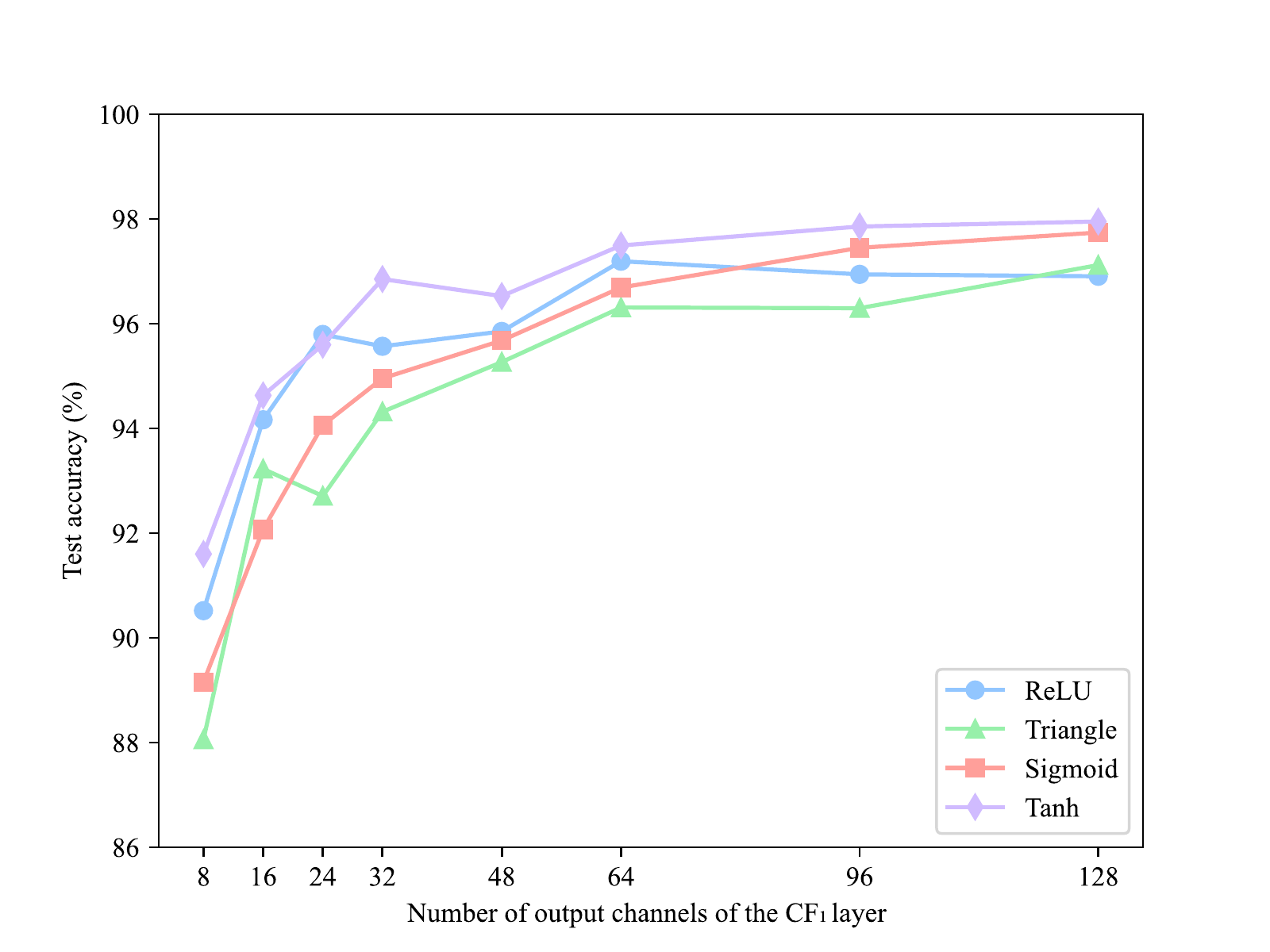}
\caption{Effect of the number of output channels and activation functions.}
\label{fig:analysis:af_oc}
\end{figure}

Our experiments consider several activation functions (which we will discuss in the next section) and the number of output channels for the CF$_1$ layer. As shown in Fig. \ref{fig:analysis:af_oc}, our methods with different activation functions generally achieve higher performance by learning more feature maps as expected.
However, as the number of output channels increases, the accuracy of all methods rapidly saturates. In other words, we can attempt to increase the width (number of output channels) of the ConvBLS  to improve accuracy, which is the basis of the incremental learning algorithm for ConvBLS. 

\subsubsection{The effect of activation functions}
The selection of the activation function is also an important issue. In this part, four activation functions, including Tanh, ReLU, Sigmoid, and Triangle, are validated.
As shown in Fig. \ref{fig:analysis:af_oc}, the Tanh function works best, while the Triangle has the worst accuracy. Besides, the ReLU and Sigmoid activation functions have similar accuracy. Given that Triangle is computationally expensive and its accuracy is the worst, we further investigate the remaining three activation functions in the following experiments.  
Lastly, considering the simplicity of ReLU, we use the ReLU activation function to construct our ConvBLS model.

\subsubsection{The effect of kernel sizes}
\begin{figure}[!t]
\centering
\includegraphics[width=2.8in]{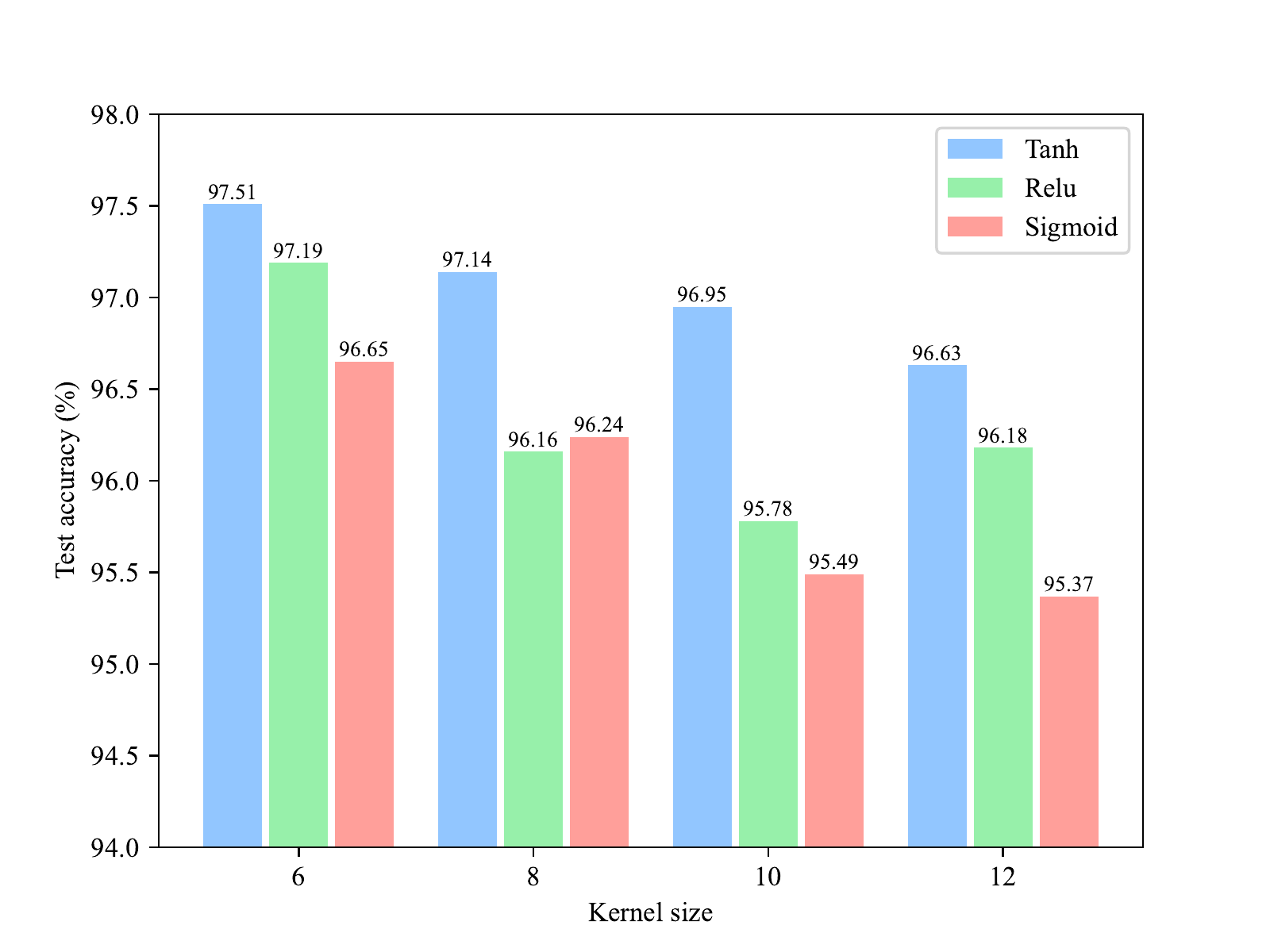}
\caption{Effect of kernel sizes and activation functions.}
\label{fig:analysis:af_k}
\end{figure}

Different kernel sizes can capture features of different scales. Hence, we test kernel sizes of $6$, $8$, $10$, and $12$. As shown in Fig. \ref{fig:analysis:af_k}, the methods with a smaller kernel size in CF layers work better. The reason for the above phenomenon is as follows. In particular, as the kernel size increases, the dimension of the feature space where the unsupervised learning algorithm works also becomes larger. This makes it difficult for the algorithm to discover the selective filters.

\subsubsection{The effect of strides}
\begin{figure}[!t]
\centering
\includegraphics[width=2.8in]{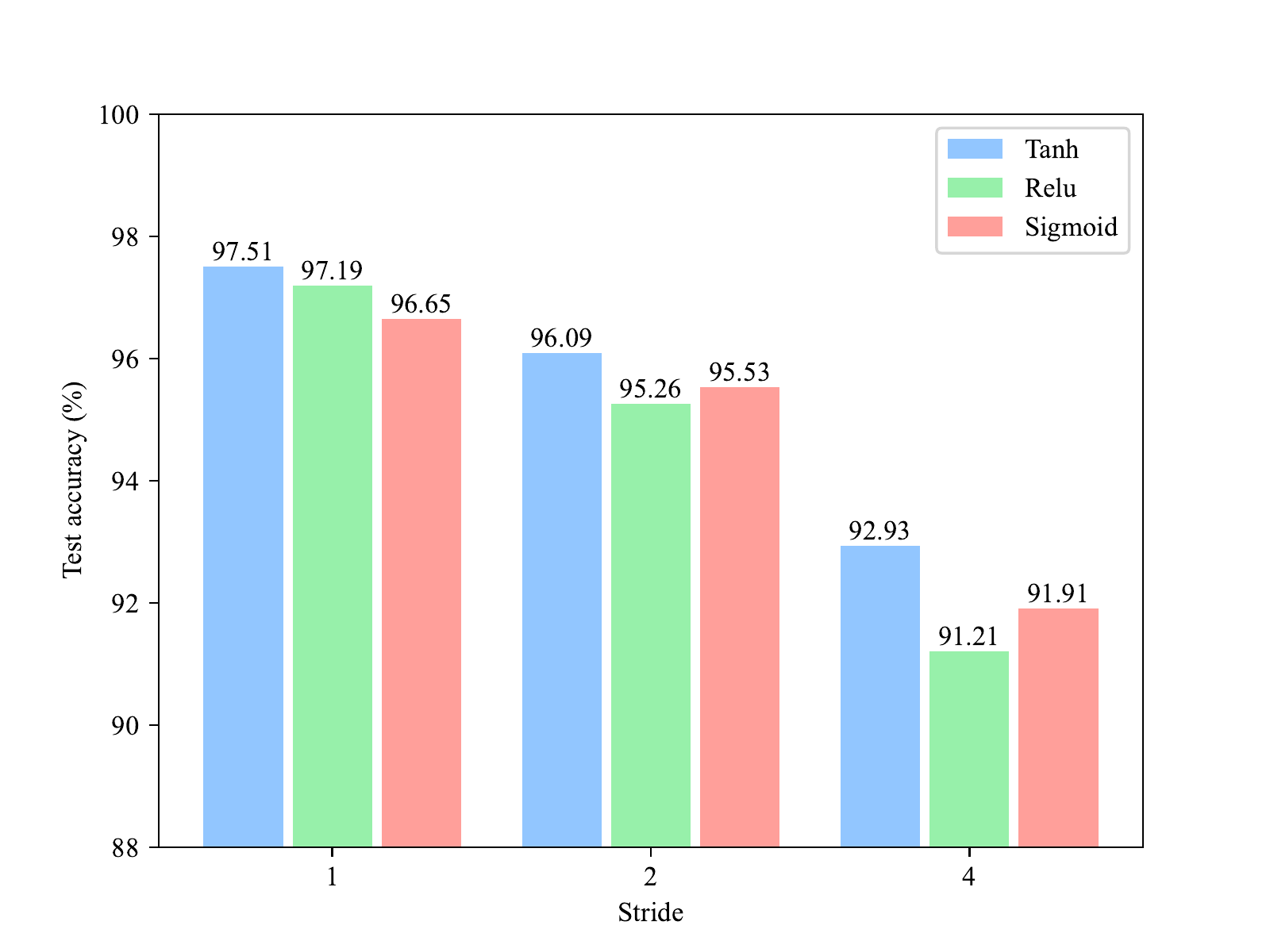}
\caption{Effect of strides and activation functions.}
\label{fig:analysis:af_s}
\end{figure}

We vary the stride over $1$, $2$, and $4$. Fig. \ref{fig:analysis:af_s} depicts the results.
Overall, the 1-pixel stride works best and the larger the stride the worse the accuracy. Because the larger the stride, the less the number of features we can extract. Thus, if we have computational resource to spare, our results suggest that it is better to spend it on reducing the stride and selecting a small kernel size.

\section{Conclusions and Future Work}
In this article, a convolutional broad learning system is proposed for image classification. To design an efficient and effective ConvBLS, we provide a solution from the perspective of the model architecture and training algorithm simultaneously. On one hand, a ConvBLS architecture is developed, which consists of the CF layer, CE layer, TSMS feature fusion layer, and output layer. Thanks to the architectural design and TSMS feature fusion mechanism, the architecture of ConvBLS is effective. On the other hand, a training algorithm for ConvBLS is proposed. Benefiting from 
the SKM for the CF layers and supervised learning for the output layer, the training of our ConvBLS is very efficient. Finally, we develop two corresponding incremental learning algorithms to adjust the structure of the model dynamically. Experiments on MNIST, Fashion-MNIST, and NORB datasets clearly demonstrate the effectiveness and efficiency of our ConvBLS.

There are still some works worthy of in-depth study. First, the proposed training algorithm for our ConvBLS is a framework about using unsupervised learning algorithm to optimize the weights of CF layers and using supervised learning algorithm to calculate the weights of the output layer. The SKM is just used as an example. Therefore, whether other unsupervised learning algorithms are more suitable for such a training framework is a question worth investigating. Second, it is necessary to further investigate how to tune the hyper-parameters and other complex architectural parameters. All hyperparameters in ConvBLS are tuned manually based on expert experience, which is inefficient. In the next step, we will try to use the neural architecture search approach for automatic tuning of hyper-parameters. At last, our ConvBLS is solely a theoretical framework, which is validated on generic datasets. In the following, we can apply the ConvBLS to specific scenario, such as human action recognition, face recognition, and facial expression recognition.

\bibliographystyle{IEEEtran}
\bibliography{./main.bib}
 
\begin{IEEEbiography}[{\includegraphics[width=1in,height=1.25in,clip,keepaspectratio]{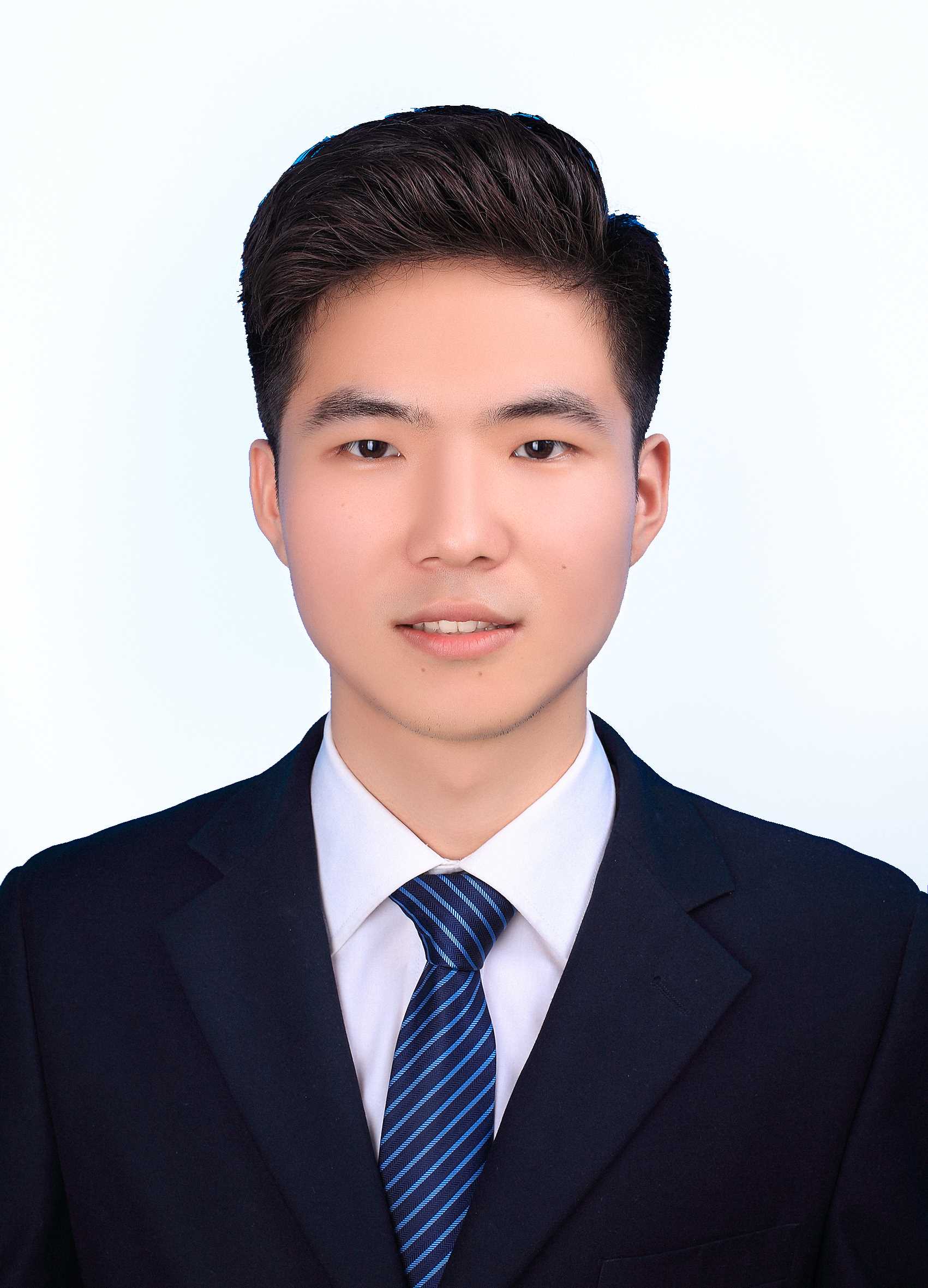}}]{Chunyu Lei}
	received the B.S. degree in computer science and technology from Zhengzhou University, Zhengzhou, China, in 2020. He is currently pursuing the Ph.D. degree in computer science and technology from South China University of Technology, Guangzhou, China. 
    
    His research interests include broad learning system, neural architecture search, and computational intelligence.
\end{IEEEbiography}

\begin{IEEEbiography}[{\includegraphics[width=1in,height=1.25in,clip,keepaspectratio]{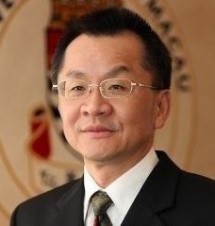}}]{C. L. Philip Chen}
	(S’88–M’88–SM’94–F’07) received the M.S. degree from the University of Michigan at Ann Arbor, Ann Arbor, MI, USA, in 1985 and the Ph.D. degree from the Purdue University in 1988, all in electrical and computer science.
	 
	He is the Chair Professor and Dean of the College of Computer Science and Engineering, South China University of Technology. He is the former Dean of the Faculty of Science and Technology. He is a Fellow of IEEE, AAAS, IAPR, CAA, and HKIE; a member of Academia Europaea (AE) and European Academy of Sciences and Arts (EASA). He received IEEE Norbert Wiener Award in 2018 for his contribution in systems and cybernetics, and machine learnings. He is also a highly cited researcher by Clarivate Analytics in 2018, 2019, 2020, 2021, and 2022. 
	
	He was the Editor-in-Chief of the IEEE Transactions on Cybernetics (2020-2021) after he completed his term as the Editor-in-Chief of the IEEE Transactions on Systems, Man, and Cybernetics: Systems (2014-2019), followed by serving as the IEEE Systems, Man, and Cybernetics Society President from 2012 to 2013. Currently, he serves as a deputy director of CAAI Transactions on Artificial Intelligence, an Associate Editor of the IEEE Transactions on Artificial Intelligence, IEEE Trans on SMC: Systems, and IEEE Transactions on Fuzzy Systems, an Associate Editor of China Sciences: Information Sciences. He received Macau FDCT Natural Science Award three times and a First-rank Guangdong Province Scientific and Technology Advancement Award in 2019. His current research interests include cybernetics, computational intelligence, and systems.
\end{IEEEbiography}

\begin{IEEEbiography}[{\includegraphics[width=1in,height=1.25in,clip,keepaspectratio]{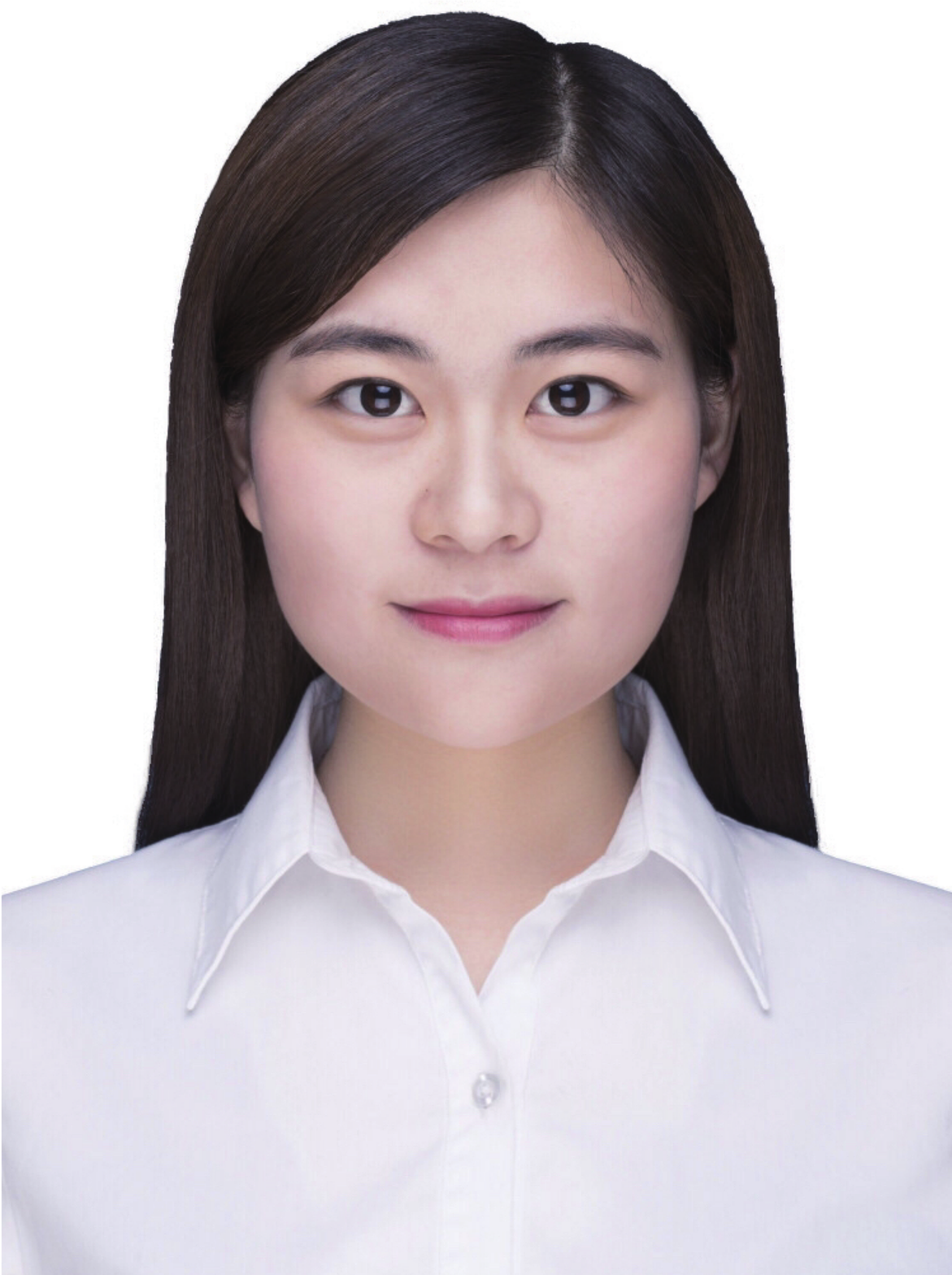}}]{Jifeng Guo}
	received the B.S. and M.S. degrees in computer science and technology from the University of Jinan, Jinan, China, in 2016 and 2019, respectively. She is currently pursuing the Ph.D. degree in computer science and technology with the South China University of Technology, Guangzhou, China.
	
    Her current research interests include computational intelligence, semi-supervised learning, broad learning systems, deep learning, emotion recognition, computer simulation, cement modeling, and data mining.
\end{IEEEbiography}

\begin{IEEEbiography}[{\includegraphics[width=1in,height=1.25in,clip,keepaspectratio]{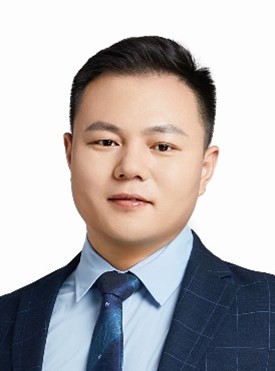}}]{Tong Zhang}
	(S'12-M'16) received the B.S. degree in software engineering from Sun Yat-sen University, at Guangzhou, China, in 2009, and the M.S. degree in applied mathematics from University of Macau, at Macau, China, in 2011, and the Ph.D. degree in software engineering from the University of Macau, at Macau, China in 2016. Dr. Zhang currently is a professor with the School of Computer Science and Engineering, South China University of Technology, China.
	
	His research interests include affective computing, evolutionary computation, neural network, and other machine learning techniques and their applications. Dr. Zhang is an Associate Editor of the IEEE Transactions on Computational Social Systems. He has been working in publication matters for many IEEE conferences.
\end{IEEEbiography}

\end{document}